\documentclass{article}
\pdfoutput=1 %PER ARXIV
\usepackage{amssymb}

\usepackage{listings}

\usepackage{physics}
\usepackage{amsmath}
\usepackage{tikz}
\usepackage{mathdots}
\usepackage{cancel}
\usepackage{color}
\usepackage{siunitx}
\usepackage{array}
\usepackage{multirow}
\usepackage{amssymb}
\usepackage{gensymb}
\usepackage{tabularx}
\usepackage{extarrows}
\usepackage{booktabs}
\usetikzlibrary{fadings}
\usetikzlibrary{patterns}
\usetikzlibrary{shadows.blur}
\usetikzlibrary{shapes}
\usepackage{arydshln}
\usepackage[normalem]{ulem}
\usepackage{amsmath}
\usepackage{graphicx,xcolor}
\usepackage{placeins}

\usepackage{subcaption}
\usepackage{booktabs}

\usepackage{tikz}
\usetikzlibrary{positioning, shapes, shadows, arrows}
\usetikzlibrary{decorations.pathreplacing}
\tikzstyle{abstract}=[circle, draw=black, fill=white]
\tikzstyle{labelnode}=[circle, draw=white,opacity=.2,text opacity=1]
\tikzstyle{invisiblenode}=[circle,dashed, inner sep=1pt,circle split,line width=1mm,minimum size=1.5cm]
\tikzstyle{line} = [draw, -latex']
\newcommand{\subf}[2]{%
  {\small\begin{tabular}[t]{@{}c@{}}
  #1\\#2
  \end{tabular}}
}

\usepackage[utf8]{inputenc}

\title{Toward the application of XAI methods in EEG-based systems}
\author{Andrea Apicella, Francesco Isgrò, Andrea Pollastro, Roberto Prevete\\
\textit{\small Laboratory of Augmented Reality for Health Monitoring (ARHeMLab)}\\
\textit{\small Laboratory of Artificial Intelligence, Privacy \& Applications (AIPA Lab)}\\
\textit{\small Department of Electrical Engineering and Information Technology, University of Naples Federico II}}
\date{}

\usepackage[numbers]{natbib} %[NUMBERS] IMPORTANTE PER ARXIV
\usepackage{graphicx}

\begin{document}

\maketitle

\begin{abstract}
\footnote{accepted to be presented at \textit{XAI.it 2022 - Italian Workshop on Explainable Artificial Intelligence}. Please refer to the published version at \url{https://ceur-ws.org/Vol-3277/paper1.pdf}}
    An interesting case of the well-known Dataset Shift Problem is the classification of Electroencephalogram (EEG) signals in the context of Brain-Computer Interface (BCI). The non-stationarity of EEG signals can lead to poor generalisation performance in BCI classification systems used in different sessions, also from the same subject. In this paper, we start from the hypothesis that the Dataset Shift problem can be alleviated by exploiting suitable eXplainable Artificial Intelligence (XAI) methods to locate and transform the relevant characteristics of the input for the goal of classification. In particular, we focus on an experimental analysis of explanations produced by several XAI methods on an ML system trained on a typical EEG dataset for emotion recognition. Results show that many relevant components found by XAI methods are shared across the sessions and can be used to build a system able to generalise better. However, relevant components of the input signal also appear to be highly dependent on the input itself.

\end{abstract}

In this research work, we experimentally investigate the performances of several well-known eXplainable Artificial (XAI) methods proposed in the literature in the context of Brain-Computer Interface (BCI) problems using EEG input-based Machine Learning (ML) algorithms to evaluate the possibility of alleviating the \textit{Dataset Shift problem}. This is not a trivial issue as, differently from other signals, the non-stationarity of EEG signals makes them hard to analyse.
In recent years, Brain-Computer Interfaces (BCIs) have been emerging as technology allowing the human brain to communicate with external devices without the use of peripheral nerves and muscles, enhancing the interaction capability of the user with the environment. In particular, several proposals of BCI methods based on Electroencephalographic (EEG) signals are receiving growing interest by the scientific community thanks to its  implication in medical purposes \cite{apicella2021high,arpaia2021wearable}, other than other fields such as  entertainment \cite{marshall2013games}, education \cite{apicella2022eeg}, and marketing \cite{mashrur2022bci}. This is because measuring and monitoring the brain's electrical activity can provide important information related to the brain's physiological, functional, and pathological status. EEG signals are particularly suitable to this aim thanks to their important qualities such as non-invasiveness and high temporal resolution \cite{subha2010eeg}. Furthermore, several solutions for comfortable and wearable EEG acquisition devices are being proposed \cite{casson2010wearable,arpaia2021metrological}, allowing an acquisition process less influenced by noise due to the user-device interaction. Thanks to its properties, the EEG signal is one of the most promising candidates to become one of the most used communication channels between man and machine. 

Several BCI solutions adopting ML methods are proposed in the literature. Generally, EEG data acquired from persons subjected to well-known stimuli are used in the training stage. These data are labelled following some established protocol, usually dependent on the task. For example, in an Emotion Recognition (ER) task, stimuli can be images or videos considered able to elicit particular emotions. Therefore the labels can be inferred by the stimuli or declared by the subject, who will say whether or not he felt a specific emotion during the stimulus administration. If the training stage is successful, the model can generalise on new unlabelled data, such as new acquisition from another subject or the same subject in another session.

However, one of the main defects of the EEG signal is that its statistical characteristics change over time. This implies that even under the same conditions and for the same task, significantly different signals can be acquired just as time passes. It is important to highlight that this phenomenon can also occur using the same stimuli-reaction (e.g., same emotions with the same stimuli) to the same subject at different times, leading to substantially different EEG signals even for the same subject. This problem is even more present among different subjects, who, given the same stimuli and emotions, can produce very different acquisitions between them. For these reasons, EEG is considered a non-stationary signal \cite{kaplan2005nonstationary}.
%This implies that, even under the same conditions and for the same task, significantly different signals can be acquired just as time passes. It is important to highlight that this phenomenon can also occur using the same stimuli-reaction (e.g., same emotions with the same stimuli in an emotion detection task) to the same subject at different times, leading to substantially different EEG signals even for the same subject. This problem is obviously even more highlighted among different subjects who, given the same stimuli and emotions, can give rise to very different acquisitions between them. For these reasons, EEG is considered a non-stationary signal \cite{kaplan2005nonstationary}. %Because of this, strong differences across acquisitions made in different times or across different subjects can arise, even with the same stimuli and reactions, leading to different probability distributions of the acquired data. 
More in detail, the following cases in an EEG-based task can arise: 
i) a model trained on a set of EEG data acquired from a given subject at a specific time could not work on data acquired from the same subject at different times (Cross-Session generalisation problem), or  
ii) a model trained on data acquired from one or more subjects should not work as expected in classifying EEG signals acquired from a different subject at different times (Cross-Subject generalisation problem). 

This type of problem can be treated as an instance of the Dataset Shift problem \cite{quinonero2008dataset}. In a nutshell, Dataset Shift arises when the distribution of the training data differs from the data distribution used outside of the training stage (that is, running or evaluation stages); therefore the standard ML assumption \cite{quinonero2008dataset} to have the same data distribution for both training and test set does not hold. 
Consequently, standard ML approaches can produce ML systems which exhibit poor generalisation performances.

%Notice that the Cross-Subject generalisation problem could be mitigated by training specific models for each subject (Subject-Dependent models), reducing the performance gap due to using the same ML system trained on different users. However, a Subject-Dependent model can be considered effective only on data acquired by the same training subject, resulting in an expensive and not very versatile BCI system, as well as uncomfortable to the user due to initial acquisition sessions before the effective use of the system.

%\rob{FORSE QUESTO PEZZO LO TOGLEREI, VALUTIAMO: Summarising, applying ML models in the pipeline of EEG-based systems can result in poor generalisation due to the intrinsic characteristics of the EEG signal. Indeed, the highly non-stationary of the EEG signals makes the distribution of the training data different from the data distribution used outside of the training stage (that is, running or evaluation stages), making the dataset shift problem very much felt in this context.}

\tikzset{every picture/.style={line width=0.75pt}} %set default line width to 0.75pt        
\begin{figure*}
\begin{center}
\tikzset{every picture/.style={line width=0.75pt}} %set default line width to 0.75pt        

\tikzset{every picture/.style={line width=0.75pt}} %set default line width to 0.75pt        

\begin{tikzpicture}[x=0.75pt,y=0.75pt,yscale=-1,xscale=1]
%uncomment if require: \path (0,391); %set diagram left start at 0, and has height of 391

%Shape: Rectangle [id:dp04955892326530731] 
\draw   (241,94) -- (346.5,94) -- (346.5,167) -- (241,167) -- cycle ;
%Straight Lines [id:da9378213481031472] 
\draw    (204.5,126) -- (239.5,126) ;
\draw [shift={(241.5,126)}, rotate = 180] [color={rgb, 255:red, 0; green, 0; blue, 0 }  ][line width=0.75]    (10.93,-3.29) .. controls (6.95,-1.4) and (3.31,-0.3) .. (0,0) .. controls (3.31,0.3) and (6.95,1.4) .. (10.93,3.29)   ;
%Straight Lines [id:da55302271271784] 
\draw    (346.5,129) -- (433.5,128.02) ;
\draw [shift={(435.5,128)}, rotate = 179.36] [color={rgb, 255:red, 0; green, 0; blue, 0 }  ][line width=0.75]    (10.93,-3.29) .. controls (6.95,-1.4) and (3.31,-0.3) .. (0,0) .. controls (3.31,0.3) and (6.95,1.4) .. (10.93,3.29)   ;
%Straight Lines [id:da12752972153504016] 
\draw  [dash pattern={on 0.84pt off 2.51pt}]  (293.5,52) -- (294.45,94) ;
\draw [shift={(294.5,96)}, rotate = 268.7] [color={rgb, 255:red, 0; green, 0; blue, 0 }  ][line width=0.75]    (10.93,-3.29) .. controls (6.95,-1.4) and (3.31,-0.3) .. (0,0) .. controls (3.31,0.3) and (6.95,1.4) .. (10.93,3.29)   ;
%Curve Lines [id:da1113555066256886] 
\draw    (391,128.5) .. controls (392,83.72) and (408.34,18.65) .. (332.65,21.95) ;
\draw [shift={(331.5,22)}, rotate = 357.03] [color={rgb, 255:red, 0; green, 0; blue, 0 }  ][line width=0.75]    (10.93,-3.29) .. controls (6.95,-1.4) and (3.31,-0.3) .. (0,0) .. controls (3.31,0.3) and (6.95,1.4) .. (10.93,3.29)   ;
%Shape: Rectangle [id:dp8904989093048378] 
\draw   (262,5) -- (332,5) -- (332,45) -- (262,45) -- cycle ;

%Shape: Square [id:dp14298702005107444] 
\draw   (167,70) -- (188,70) -- (188,91) -- (167,91) -- cycle ;
%Shape: Square [id:dp8016080911778298] 
\draw   (167,91) -- (188,91) -- (188,112) -- (167,112) -- cycle ;
%Shape: Square [id:dp7182688756939402] 
\draw   (167,112) -- (188,112) -- (188,133) -- (167,133) -- cycle ;
%Shape: Square [id:dp9953198603440064] 
\draw   (167,133) -- (188,133) -- (188,154) -- (167,154) -- cycle ;
%Shape: Square [id:dp2974621042764002] 
\draw   (167,154) -- (188,154) -- (188,175) -- (167,175) -- cycle ;
%Shape: Square [id:dp69598930803713] 
\draw   (167,175) -- (188,175) -- (188,196) -- (167,196) -- cycle ;

%Shape: Square [id:dp472983393564547] 
\draw  [fill={rgb, 255:red, 255; green, 255; blue, 255 }  ,fill opacity=1 ] (154,75) -- (175,75) -- (175,96) -- (154,96) -- cycle ;
%Shape: Square [id:dp014380865196633397] 
\draw  [fill={rgb, 255:red, 255; green, 255; blue, 255 }  ,fill opacity=1 ] (154,96) -- (175,96) -- (175,117) -- (154,117) -- cycle ;
%Shape: Square [id:dp07372733208927917] 
\draw  [fill={rgb, 255:red, 255; green, 255; blue, 255 }  ,fill opacity=1 ] (154,117) -- (175,117) -- (175,138) -- (154,138) -- cycle ;
%Shape: Square [id:dp6224955926296536] 
\draw  [fill={rgb, 255:red, 255; green, 255; blue, 255 }  ,fill opacity=1 ] (154,138) -- (175,138) -- (175,159) -- (154,159) -- cycle ;
%Shape: Square [id:dp35320856080969276] 
\draw  [fill={rgb, 255:red, 255; green, 255; blue, 255 }  ,fill opacity=1 ] (154,159) -- (175,159) -- (175,180) -- (154,180) -- cycle ;
%Shape: Square [id:dp23303973343584483] 
\draw  [fill={rgb, 255:red, 255; green, 255; blue, 255 }  ,fill opacity=1 ] (154,180) -- (175,180) -- (175,201) -- (154,201) -- cycle ;

%Shape: Wave [id:dp884074747621377] 
\draw   (123.99,264.78) .. controls (120.14,254.44) and (116.46,244.6) .. (111.94,244.5) .. controls (107.42,244.4) and (103.3,254.07) .. (99,264.22) .. controls (94.69,274.37) and (90.58,284.04) .. (86.06,283.94) .. controls (81.53,283.84) and (77.85,274) .. (74,263.67) .. controls (70.16,253.33) and (66.47,243.5) .. (61.95,243.4) .. controls (59.29,243.34) and (56.77,246.66) .. (54.27,251.47) ;
%Shape: Wave [id:dp04760508832672705] 
\draw   (51,264) .. controls (55.08,274.25) and (58.98,284) .. (63.5,284) .. controls (68.02,284) and (71.92,274.25) .. (76,264) .. controls (80.08,253.75) and (83.98,244) .. (88.5,244) .. controls (93.02,244) and (96.92,253.75) .. (101,264) .. controls (105.08,274.25) and (108.98,284) .. (113.5,284) .. controls (116.16,284) and (118.61,280.62) .. (121,275.76) ;
%Shape: Rectangle [id:dp0649011935213526] 
\draw   (147,249) -- (217,249) -- (217,289) -- (147,289) -- cycle ;
%Straight Lines [id:da030180745606353643] 
\draw    (124.5,270) -- (144.5,270.91) ;
\draw [shift={(146.5,271)}, rotate = 182.6] [color={rgb, 255:red, 0; green, 0; blue, 0 }  ][line width=0.75]    (10.93,-3.29) .. controls (6.95,-1.4) and (3.31,-0.3) .. (0,0) .. controls (3.31,0.3) and (6.95,1.4) .. (10.93,3.29)   ;
%Straight Lines [id:da2214103130302002] 
\draw    (182.5,248) -- (182.02,198) ;
\draw [shift={(182,196)}, rotate = 89.45] [color={rgb, 255:red, 0; green, 0; blue, 0 }  ][line width=0.75]    (10.93,-3.29) .. controls (6.95,-1.4) and (3.31,-0.3) .. (0,0) .. controls (3.31,0.3) and (6.95,1.4) .. (10.93,3.29)   ;
%Curve Lines [id:da2835721647637133] 
\draw    (262.5,23) .. controls (160.53,20.03) and (157.55,23.92) .. (162.35,74.45) ;
\draw [shift={(162.5,76)}, rotate = 264.51] [color={rgb, 255:red, 0; green, 0; blue, 0 }  ][line width=0.75]    (10.93,-3.29) .. controls (6.95,-1.4) and (3.31,-0.3) .. (0,0) .. controls (3.31,0.3) and (6.95,1.4) .. (10.93,3.29)   ;
%Straight Lines [id:da5054669619534397] 
\draw    (136.5,128) -- (100.5,128) ;
\draw [shift={(98.5,128)}, rotate = 360] [color={rgb, 255:red, 0; green, 0; blue, 0 }  ][line width=0.75]    (10.93,-3.29) .. controls (6.95,-1.4) and (3.31,-0.3) .. (0,0) .. controls (3.31,0.3) and (6.95,1.4) .. (10.93,3.29)   ;
%Shape: Rectangle [id:dp9126375188504865] 
\draw  [dash pattern={on 4.5pt off 4.5pt}] (137,64) -- (203.5,64) -- (203.5,205) -- (137,205) -- cycle ;
%Shape: Axis 2D [id:dp483320682485628] 
\draw  (44,159.9) -- (117.5,159.9)(51.35,87) -- (51.35,168) (110.5,154.9) -- (117.5,159.9) -- (110.5,164.9) (46.35,94) -- (51.35,87) -- (56.35,94)  ;
%Curve Lines [id:da2215681000292531] 
\draw [color={rgb, 255:red, 0; green, 66; blue, 255 }  ,draw opacity=1 ]   (52.5,114) .. controls (66.5,116) and (35.5,167) .. (91.5,157) ;

% Text Node
\draw (348.5,127) node [anchor=north west][inner sep=0.75pt]   [align=left] {prediction};
% Text Node
\draw (252,121) node [anchor=north west][inner sep=0.75pt]  [font=\large] [align=left] {\begin{minipage}[lt]{61.9pt}\setlength\topsep{0pt}
\begin{center}
ML system
\end{center}

\end{minipage}};
% Text Node
\draw (265,60) node [anchor=north west][inner sep=0.75pt]   [align=left] {examines};
% Text Node
\draw (268,7) node [anchor=north west][inner sep=0.75pt]  [xslant=-0.02] [align=left] {\\{\large  \ \ \ XAI}\\ {\large method}};
% Text Node
\draw (70,282) node [anchor=north west][inner sep=0.75pt]   [align=left] {EEG};
% Text Node
\draw (137.39,188.1) node [anchor=north west][inner sep=0.75pt]  [rotate=-269.33] [align=left] {relevance};
% Text Node
\draw (150,247) node [anchor=north west][inner sep=0.75pt]   [align=left] {\\{\large Feature}\\{\large extractor}};
% Text Node
\draw (186.31,182.73) node [anchor=north west][inner sep=0.75pt]  [rotate=-268.35] [align=left] {features};
% Text Node
\draw (14,176) node [anchor=north west][inner sep=0.75pt]   [align=left] {feature evaluation};
\end{tikzpicture}

\end{center}
\caption{A general functional scheme of a Machine Learning (ML) architecture based on XAI methods to select and transform relevant input features with the aim of improving the performance of ML systems in the context of the dataset-shift problem.}
\label{img:xai}
\end{figure*}
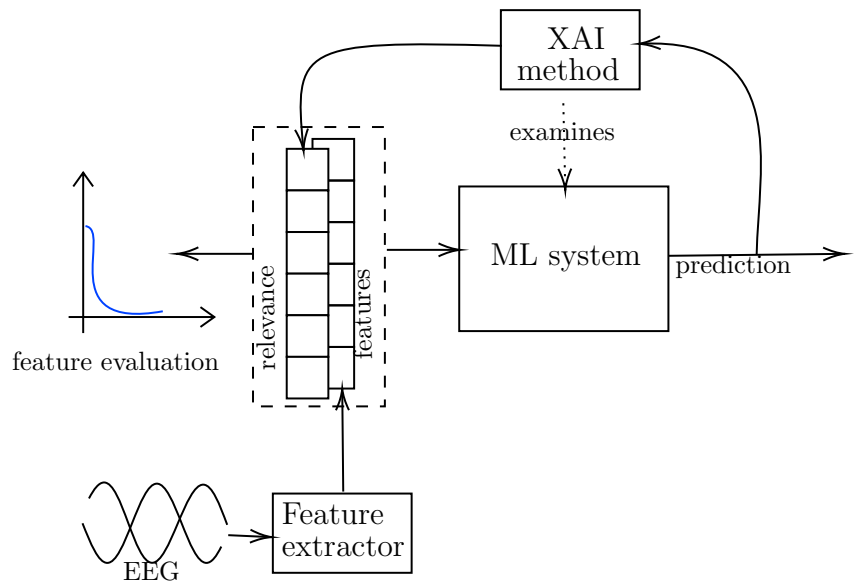

On another side, a sub-field of Artificial Intelligence, eXplainable Artificial Intelligence (XAI), wants to explain the behaviour of AI systems, such as ML ones. In general, an explanation gives information on why an ML model returns an output given a specific input. In particular, several XAI methods applied to Deep Neural Networks are giving promising results \cite{apicella2019contrastive,montavon2019layer,ribeiro2016should,apicella2015sparse}.

In the XAI context, several explanations are built by inspecting the model's inner mechanism to understand the input-output relationships, assigning a relevance score to each input component. 
However, building an explanation is particularly challenging if the model to inspect is a DNN; this is mainly for two reasons: i) DNNs offer excellent performances in several tasks, but at the price of high inner complexity of the models, leading toward low interpretability, ii) to help the ML user to understand the system behaviours, typical explanations have to be humanly understandable. 
 
The general idea of this work is that outputs' explanations of a trained ML model on given inputs can help the setup of new models able to overcome/mitigate the dataset shift problem, in general, and to generalise across subjects/sessions in case of EEG signals, in particular. 
%Therefore, explanations of an ML system can be used to locate the main characteristics of the input for each given output, and then exploited to build a new ML system able to generalise toward different datasets coming from different probability distributions. In particular, we plan to develop solutions in the context of EEG signal classification problems which can lead to mitigate the dataset shift problem due to complexity of the EEG signal. However, in our knowledge, no study in the current literature analyse classical XAI methods in the EEG domain. 

More specifically, in this work, we focus on how several well-known XAI methods proposed in literature behave in explaining decisions made by an ML system based on EEG input features (Fig. \ref{img:xai}). Notice that several current XAI methods are usually tested on datasets, such as image and text recognition datasets \cite{ribeiro2016should,apicella2022exploiting}, where the domain shift problem is slight or not present. 
%Indeed, differently from other signals, the non-stationary of the EEG signal make it hard to be analysed since its statistical properties changing continuously over the time. Therefore, highly different signals can arise also from the same subject stimuli in different sessions. leading classical XAI methods to not necessarily perform well on such complex signals. 
%In this work, several XAI methods are used to explore the generalisation performance of a ML model and understanding the relations built by the ML model between input and output.
Therefore, this work is a first step toward a long term goal consisting in exploiting explanations made by XAI methods to locate and transform the main characteristics of the input for each given output, and to build ML systems able to generalise toward different data coming from different probability distributions (in this context, sessions and subjects).
%We believe that the first steps of this approach should be the following: 1) identifying the appropriate XAI methods in the specific domain, in this case EEG classification, and 2) Evaluating which input features identified by the XAI method can be effectively used for the classification phase in a robust way with respect the non-stationarity of the EEG signal. 
To this end, in this paper, we evaluate and analyse the explanations produced by a set of well-known XAI methods on an ML system trained on data taken from SEED \cite{zheng2015investigating}, a public EEG dataset for an emotion classification task. The results obtained show, on oneside, that only some well-known XAI methods produce reliable explanations in the EEG domain in the analysed task. On another side, it is shown that the relevant components found in the training data can only be partially used on data acquired outside of the training stage. 
%This is in line with the difficulties highlighted in the current literature in producing EEG-based BCI systems able to generalise toward different sessions/subjects, thus requiring further studies and investigations to overcome the dataset shift problem in the EEG classification domain.
Notably, many relevant components found in the training data are still relevant across the sessions.

The paper is organised as follows:
In Section \ref{sec:related}, a brief description of the related works is reported. In Section \ref{sec:methods} the proposed evaluation framework is presented. In Section \ref{sec:results} the obtained results are discussed. Finally, in Section \ref{sec:conclusions} is devoted to final remarks and future
developments.

\section{Related works}
\label{sec:related}
In general, Modern ML approaches, as Deep learning, are characterised by a lack of transparency of their internal mechanisms, making it not easy for the AI scientist to understand the real reasons behind the inner behaviours. In this case, the relationships of the classified emotion with the EEG input are often challenging to understand. In the EEG-based applications, works based on simple features selection strategies to choose the best EEG features are widely proposed in the literature, such as \cite{wosiak2020hybrid,zheng2021portable}. These studies, however, are based on standard feature selection methods, without exploiting information given by XAI methods.
XAI is a branch of AI concerned to “explain” ML behaviours. This is made providing methods for generating possible explanations of the model’s outputs. XAI methods are gaining prominence in explaining several classification systems based on several inputs, such as images \cite{ribeiro2016should,apicella2019explaining},  natural language processing \cite{qian2021xnlp}, clinical decision support systems \cite{schoonderwoerd2021human}, and so on. To the best of our knowledge, however, the number of research works which attempt to improve the performance of ML models on the basis of XAI's methods is enough limited, especially in the context of bio-signal classification problems. For example, in \cite{laxmi2022modeling,selvam2022explainable} feature selection procedures are carried out on biomedical data leveraging on  Correlation-based Feature Selection and  Chaotic Spider Monkey Optimization methods. 
In \cite{ieracitano2022novel} the authors propose to use an occlusion sensitivity analysis strategy \cite{zeiler2014visualizing} to locate the most relevant cortical areas in a motor imagery task. In \cite{rathod2022review} the use of XAI methods to interpret the answer of  Epilepsy Detection systems is discussed.

\section{Methods}
\label{sec:methods}
Taking in mind that we want to use the XAI method to alleviate the dataset shift problem in the BCI context, we conducted a series of experiments having the following goals: 1) testing the capability of the selected XAI methods to find relevant components for this specific signal; 2) verifying how much relevant components are dependent on the single sample of the dataset where the relevance are computed; 3) how much relevant components can be considered shared among samples of the same session, and finally 4) how much relevant components can be considered shared between samples of two different sessions, where the data shift problem is typically present. 

%In this work, a first study to evaluate if classical XAI methods are suitable to explain ML systems based on EEG signal is reported. 

%
In the remaining of this section, a brief description of the tested XAI methods is reported, followed by the used data and model descriptions. Finally experimental assessment and the evaluation strategy adopted are reported. 

\subsection{Investigated XAI Methods}
%(see Fig. \ref{img:xai})
In this work, we analyse XAI methods proposing explanations in terms of relevance of the input components on the output returned by a given classifier. More in detail, the following XAI methods are investigated: Saliency \cite{simonyan2013deep}, 
Guided Backpropagation \cite{springenberg2014striving},
Layer-wise Relevance Propagation (LRP) \cite{bach2015pixel}, Integrated Gradients \cite{sundararajan2017axiomatic}, and DeepLIFT \cite{shrikumar2017learning}. 
\subsubsection{Saliency}
Saliency method is one the of the simplest and more intuitive method to build an explanation of a ML system. Proposed in \cite{simonyan2013deep}, Saliency method is based on the gradient of the output function of the ML system respect to its input. In a nutshell, an explanation of the output $C(\mathbf{x})$  of a ML system fed with an input $\mathbf{x} \in \mathbb{R}^d$ is built generating a saliency map leveraging on the gradient $\frac{\partial C}{\partial \mathbf{x}}$ of $C$ with respect to its input computed through backpropagation. The magnitude of the gradient indicates how much the features need to be changed to affect the class score.

%starting from the hypothesis that $C(\mathbf{x})$ can be approximated by its first order Taylor expansion $C(\mathbf{x}) \simeq \frac{\partial C}{\partial \mathbf{x}}\mathbf{x}+b$.

\subsubsection{Guided BackPropagation}
Guided BackPropagation (Guided BP) \cite{springenberg2014striving} can be viewed as a slighlty variation of Saliency method proposed in \cite{simonyan2013deep}. The main difference is in the value used as gradient in case of rectified activation functions (ReLU): in Saliency method, the real gradient is used in computing the features relevance. Instead, Guided BP starts from the hypothesis that the user is not interested if a feature "decreases" (i.e., negative value) a neuron activation, but only in the most relevant ones. Therefore, instead of the true gradient, in guided BP a gradient transformation is used to prevent backward flow of negative values, avoiding to decrease the neuron activations and highlighting the most relevant features. Obviously, Guided BP can fail to highlight inputs that contribute negatively to the output due to "zero-ing" the negative values.

\subsubsection{Layer-wise Relevance Propagation}
Layer-wise Relevance Propagation (LRP) associates a relevance value to each input element (pixels in case of images) to build explanations for the ML model answer. In a nutshell, the output $C(\mathbf{x})$ of a ML system  on an input $\mathbf{x} \in \mathbb{R}^d$ is decomposed as a sum of relevances on the single features composing $\mathbf{x}$, i.e. $C(\mathbf{x})\simeq \sum\limits_{i=1}^d R_i$ where $R_i$ is a score of the local contribution of the $i$-th feature on the produced output. In particular, positive values denote positive
contributions, while negative values negative contributions. Applied to ANN, this principle can be generalised across each pair of consecutive layers $l$ and $l+1$ of a network composed of $L$ layers such that $\sum\limits_{i=1}^{q} R_i^{(l+1)} =\sum\limits_{i=1}^{q'} R_i^{(l)}$
where $q$ and $q'$ are the features of the layers $l+1$ and $l$ respectively. Since the final network output $C(\mathbf{x})$ of an ANN is the output of the $L$-th layer, it results that $C(\mathbf{x}) = \dots =\sum\limits_{i=1}^{q} R_i^{(l+1)} =\sum\limits_{i=1}^{q'} R_i^{(l)}= \dots =\sum\limits_{i=1}^d R_i$. This rule can be interpreted as a \text{conservation rule}, and leveraging on that different methods to compute the relevance have been proposed, depending on the type of features involved. In case of densely connected layers, the most known rule is the $z-rule$ \cite{bach2015pixel}, which takes care of the neuron activations of each layer to compute the final relevance of each layer.

\subsubsection{Integrated Gradients}
One of the main drawbacks of simple gradient-based method is that the gradient respect to the input should be small in the neighbourhood of the input features also for relevant ones.  
Instead of using only the gradient respect to the original input, \cite{sundararajan2017axiomatic} proposed to average all the gradients between the original input $\mathbf{x}$ and a baseline input $\mathbf{x}^{ref}$ (that is, an input s.t. $C(\mathbf{x}^{ref})$ results in a neutral prediction). In this way, if features of inputs closer to the baseline have higher gradient magnitudes, they are taken into account thanks to the average operator. More formally, the importance of each feature $x_i$ computed by Integrated Gradient (IG) is defined as:
$$IG(x_i)=(x_i-x^{ref}_i)\int_{\alpha=0}^1 \frac{\partial C\big(x^{ref}_i+\alpha (x_i-x^{ref}_i)\big)}{\partial x_i}\,d\alpha$$
In other words, IG aggregates the gradients along the intermediate inputs on the straight-line between the baseline and the input, selected as $\alpha \in [0,1]$ changes.

\subsubsection{DeepLIFT}
In \cite{shrikumar2017learning} a method consisting in assigning feature relevance scores according to the difference between the neurons activation and a reference activation (such as the baseline for Integrated Gradient method) is proposed. The authors proposed to compute for each feature a multiplier entity similar to a partial derivative, but leveraging over finite differences instead of infinitesimal ones. Each multiplier can be defined as $m_{\Delta x \Delta t}=\frac{R_{\Delta x \Delta t}}{\Delta x}$
and represents the ratio between i) the contribution $R_{\Delta x \Delta t}$ of the difference $\Delta x=x-x^{ref}$ from the reference $x^{ref}$ of each feature $x$ to the difference $\Delta t = t - t^{ref}$ between the output $t$ and the reference output $t^{ref}$, and ii) the difference $\Delta x$. Therefore, the authors proposed a set of rules to compute the features relevance based on the proposed multipliers exploiting a Back Propagation-based approach. 

\subsection{Data}
The SEED dataset consists of EEG signals recorded from 15 subjects stimulated by 15 film clips carefully chosen to induce negative, neutral and positive emotions. Each film clip has a duration of approximately 4 minutes.
Three sessions of 15 trials were collected for each subject.
EEG signals were recorded in 62 channels using the ESI Neuroscan System\footnote{https://compumedicsneuroscan.com}. 
During our experiments, we considered the pre-computed differential entropy (DE) features smoothed by linear dynamic systems (LDS) for each second, in each channel, over the following five bands: delta (1–3 Hz); theta (4–7 Hz); alpha (8–13 Hz); beta (14–30 Hz); gamma (31–50 Hz).

In this work, the relevant components of an EEG signal can be considered taking into account three different aspects of the signal: i) considering each single feature composing the input, ii) considering each single band composing the EEG signal, that are alpha, beta, theta, and delta, and iii) considering each single channel/electrode from which the input EEG signal was acquired. Cases ii) and iii) can be viewed as different aggregations of fixed features of the EEG signals. In the following of this work, we refer generically with the term  "components" where it is not necessary to specify if we are talking about features, bands or channels.

\subsection{Experimental assessment}
\setlength{\tabcolsep}{-6pt}
\begin{figure*}[p]
\scalebox{0.9}{
\begin{tabular}{ccccc}
\subf{\includegraphics[width=0.07\textwidth, ]{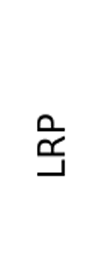}}{} & \subf{\includegraphics[width=.300\textwidth, trim={0.2cm 0.2cm 0.2cm 0.2cm},clip]{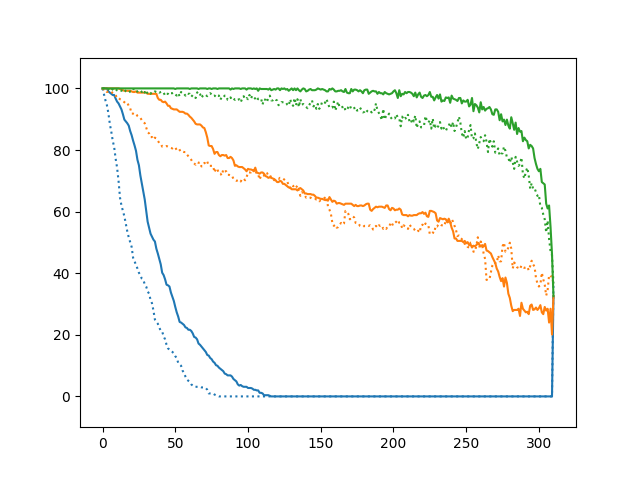}}{} & \subf{\includegraphics[width=.300\textwidth, trim={0.2cm 0.2cm 0.2cm 0.2cm},clip]{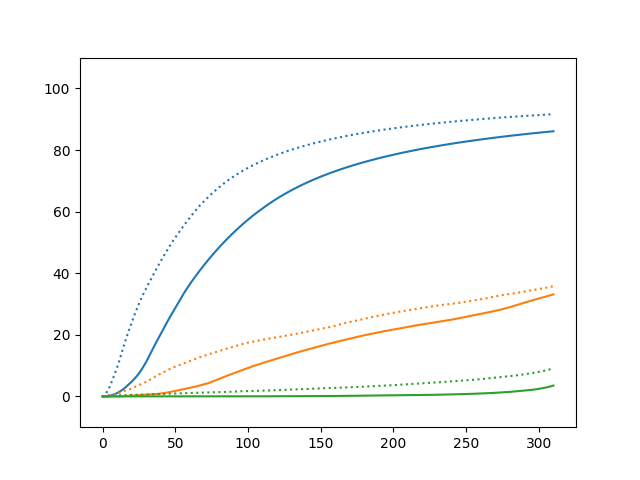}}{} & \subf{\includegraphics[width=.300\textwidth, trim={0.2cm 0.2cm 0.2cm 0.2cm},clip]{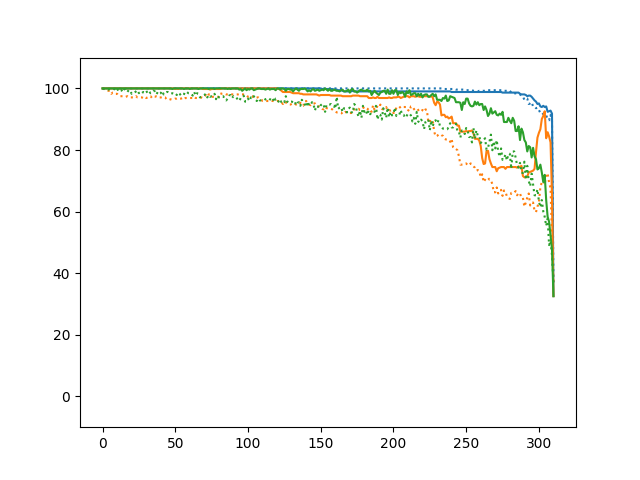}}{} & \subf{\includegraphics[width=.300\textwidth, trim={0.2cm 0.2cm 0.2cm 0.2cm},clip]{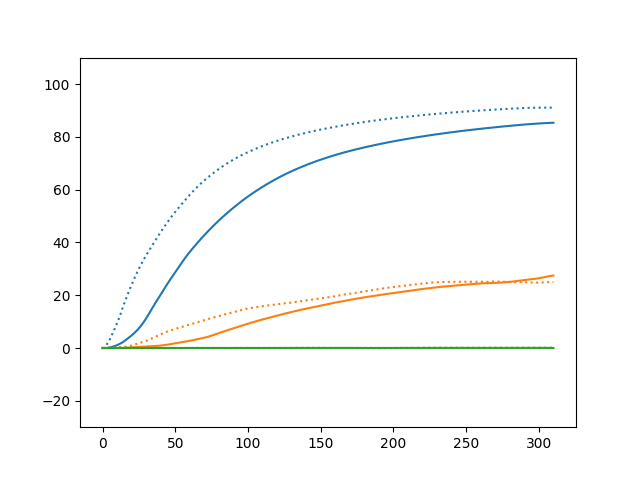}}{}\\
\subf{\includegraphics[width=0.07\textwidth]{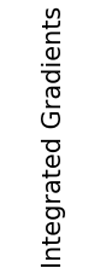}}{} & \subf{\includegraphics[width=.300\textwidth, trim={0.2cm 0.2cm 0.2cm 0.2cm},clip]{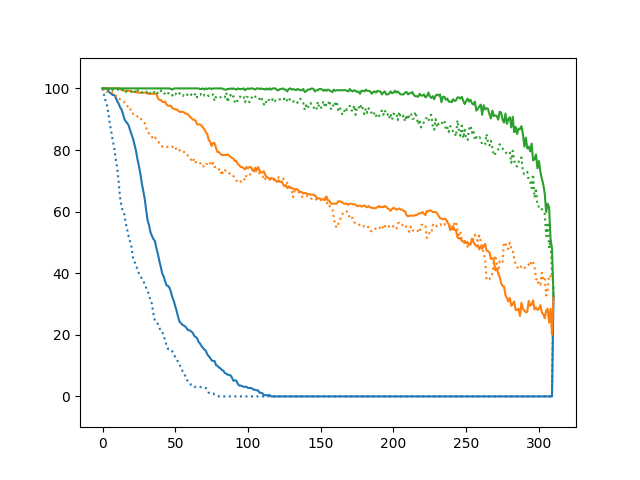}}{} & \subf{\includegraphics[width=.300\textwidth, trim={0.2cm 0.2cm 0.2cm 0.2cm},clip]{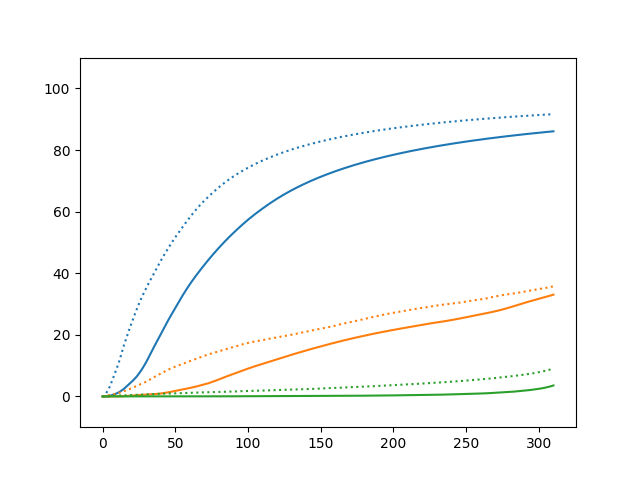}}{} & \subf{\includegraphics[width=.300\textwidth, trim={0.2cm 0.2cm 0.2cm 0.2cm},clip]{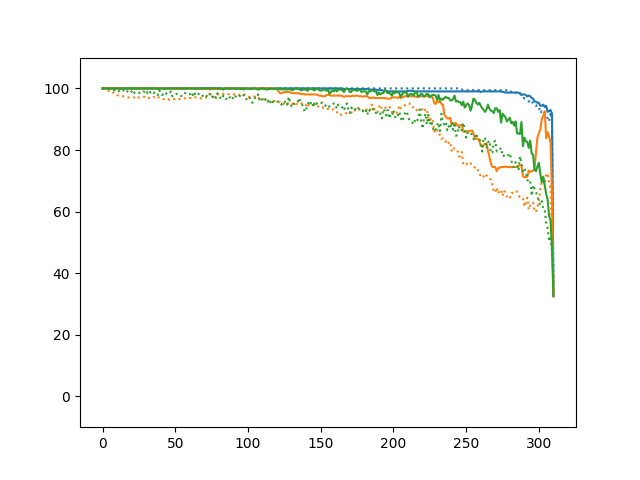}}{} & \subf{\includegraphics[width=.300\textwidth, trim={0.2cm 0.2cm 0.2cm 0.2cm},clip]{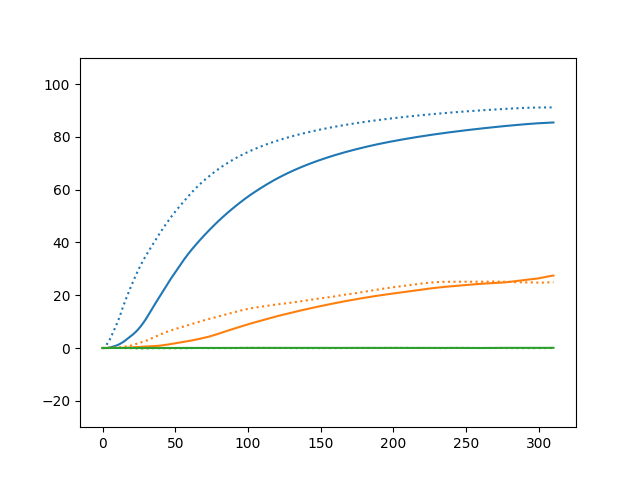}}{}\\
\subf{\includegraphics[width=0.07\textwidth]{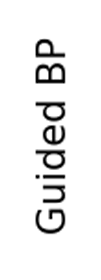}}{} & \subf{\includegraphics[width=.300\textwidth, trim={0.2cm 0.2cm 0.2cm 0.2cm},clip]{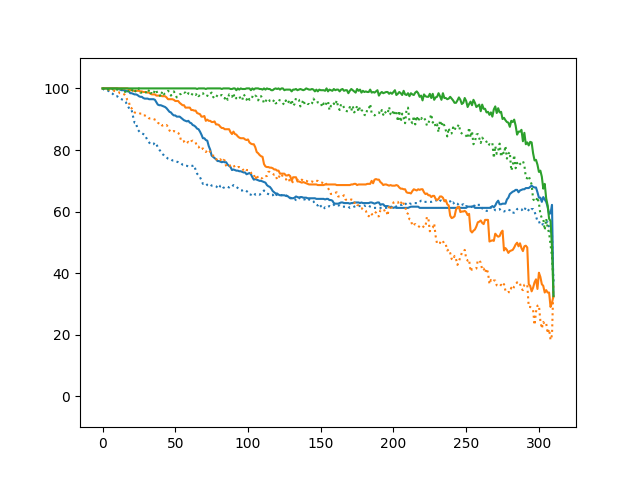}}{} & \subf{\includegraphics[width=.300\textwidth, trim={0.2cm 0.2cm 0.2cm 0.2cm},clip]{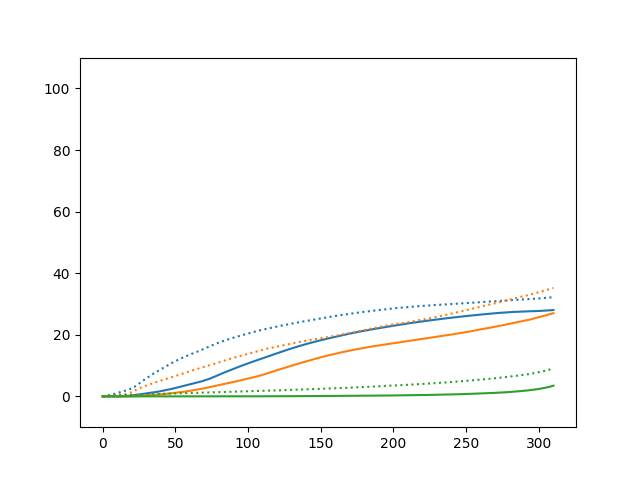}}{} & \subf{\includegraphics[width=.300\textwidth, trim={0.2cm 0.2cm 0.2cm 0.2cm},clip]{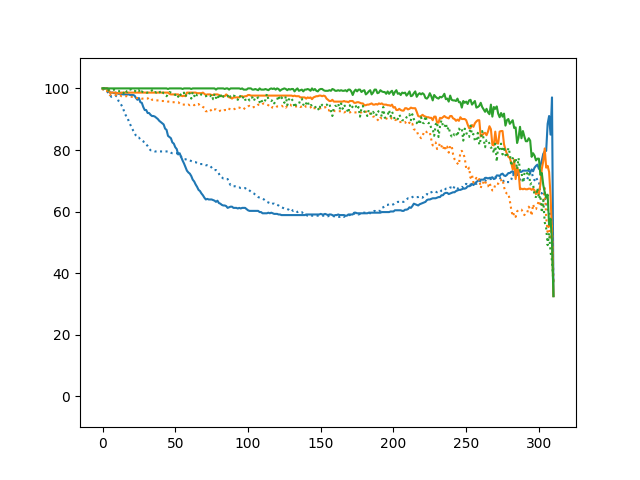}}{} & \subf{\includegraphics[width=.300\textwidth, trim={0.2cm 0.2cm 0.2cm 0.2cm},clip]{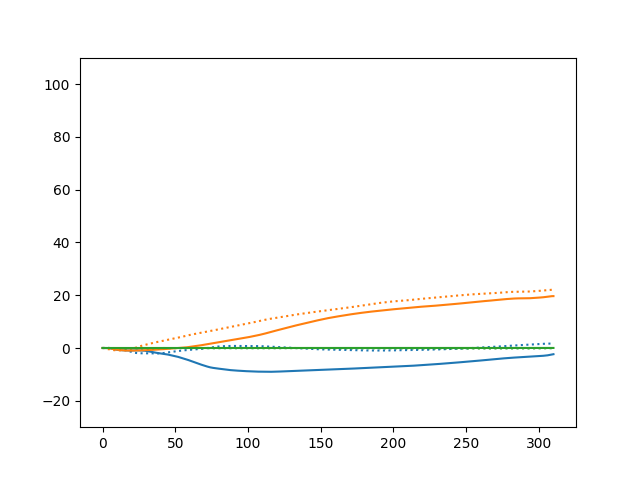}}{}\\
\subf{\includegraphics[width=0.07\textwidth]{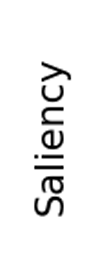}}{} & \subf{\includegraphics[width=.300\textwidth, trim={0.2cm 0.2cm 0.2cm 0.2cm},clip]{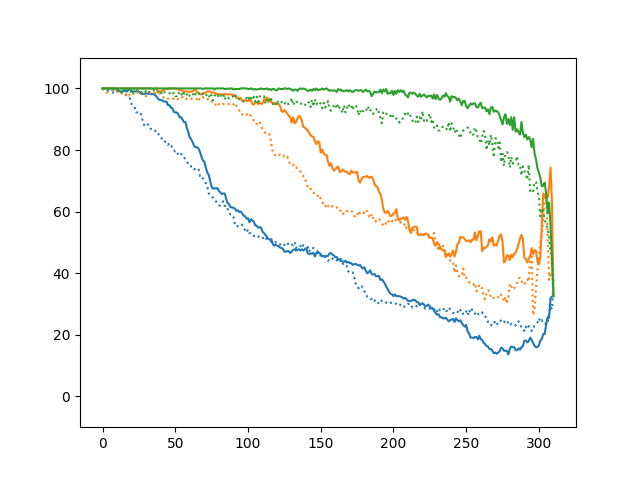}}{} & \subf{\includegraphics[width=.300\textwidth, trim={0.2cm 0.2cm 0.2cm 0.2cm},clip]{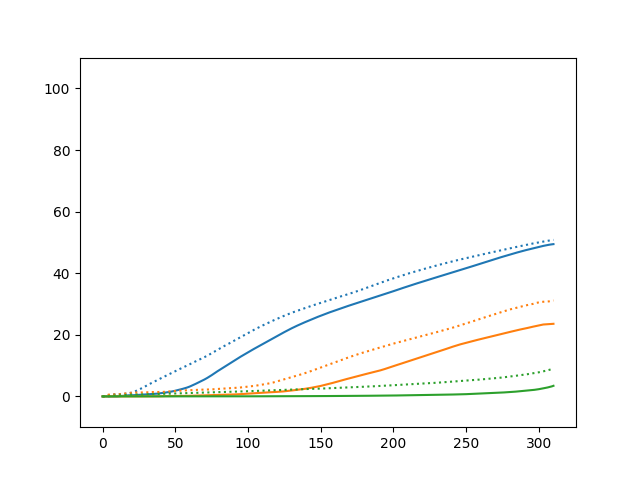}}{} & \subf{\includegraphics[width=.300\textwidth, trim={0.2cm 0.2cm 0.2cm 0.2cm},clip]{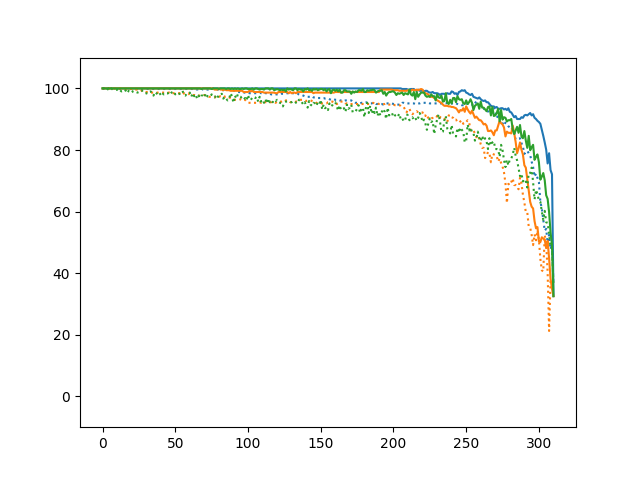}}{} & \subf{\includegraphics[width=.300\textwidth, trim={0.2cm 0.2cm 0.2cm 0.2cm},clip]{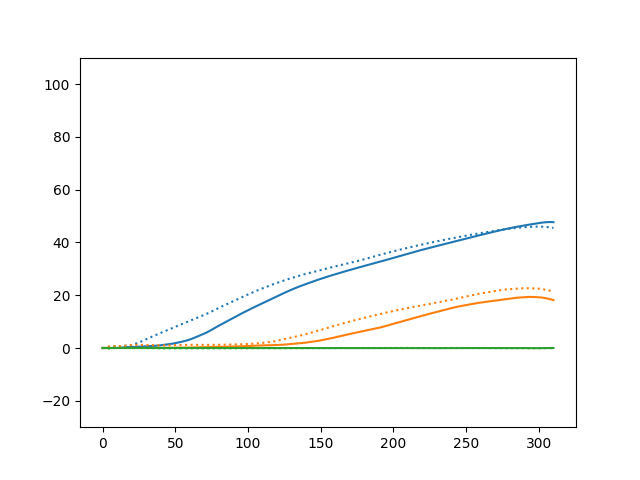}}{}\\
\subf{\includegraphics[width=0.07\textwidth]{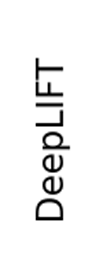}}{} & \subf{\includegraphics[width=.300\textwidth, trim={0.2cm 0.2cm 0.2cm 0.2cm},clip]{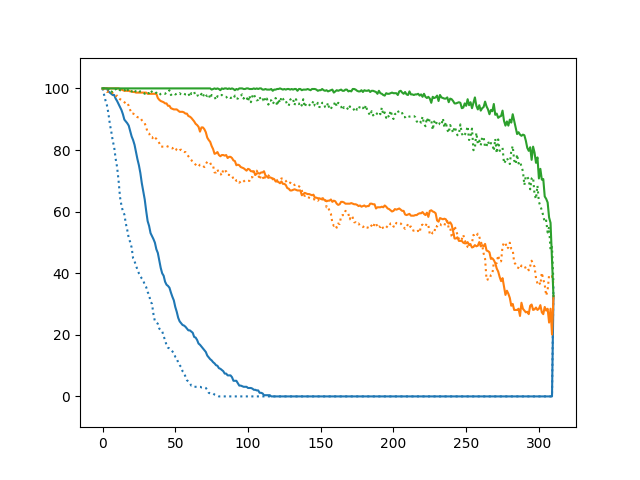}}{} & \subf{\includegraphics[width=.300\textwidth, trim={0.2cm 0.2cm 0.2cm 0.2cm},clip]{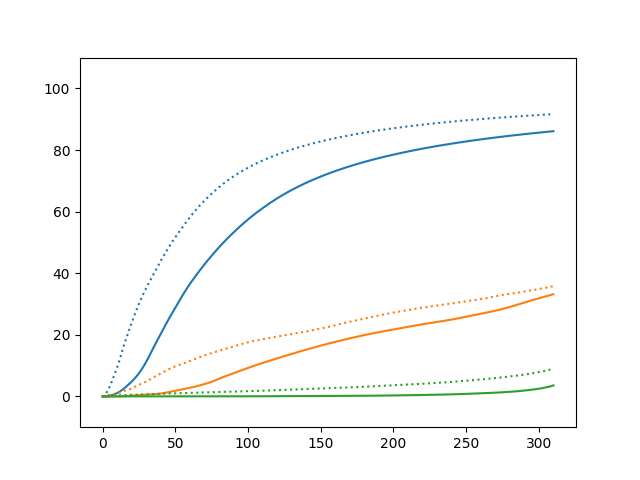}}{} & \subf{\includegraphics[width=.300\textwidth, trim={0.2cm 0.2cm 0.2cm 0.2cm},clip]{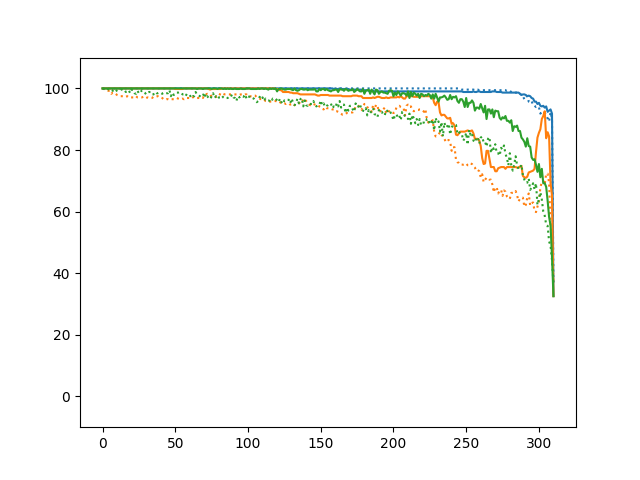}}{} & \subf{\includegraphics[width=.300\textwidth, trim={0.2cm 0.2cm 0.2cm 0.2cm},clip]{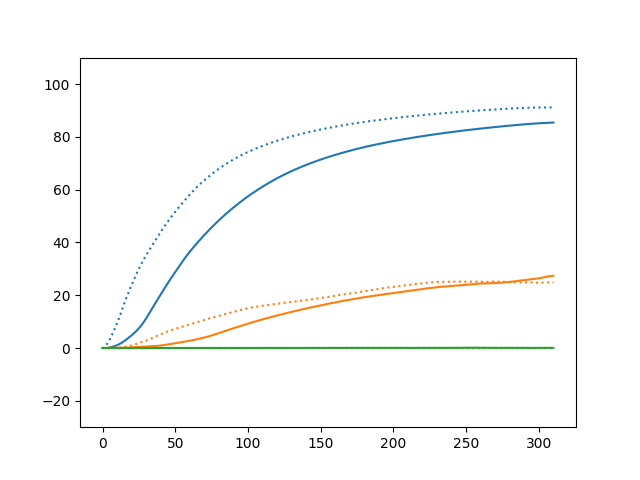}}{}\\
\hline
\end{tabular}}
\caption{MoRF (first column), AOPC (second column), LeRF (third column), and ABPC (fourth column) curves using the tested XAI methods are reported for both intra-session (solid line)  and inter-session (dotted lines) considering features as signal components. Results scoring the input components using effective relevance (blue lines) and averaged relevance computed on training data (orange lines) are reported for each case and compared with a random component scoring (green lines). On the $x$ axis and $y$ axis are reported the iteration step in the curve generation and the accuracy level reached, respectively.}
\label{img:feature}
\end{figure*}

\setlength{\tabcolsep}{-6pt}
\begin{figure*}[p]
\scalebox{0.9}{
\begin{tabular}{ccccc}
\subf{\includegraphics[width=0.07\textwidth ]{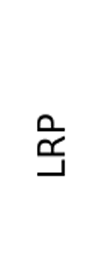}}{} & \subf{\includegraphics[width=.300\textwidth, trim={0.2cm 0.2cm 0.2cm 0.2cm},clip]{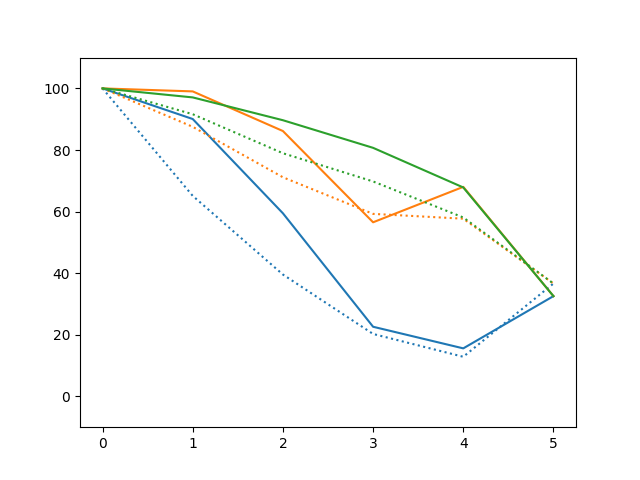}}{} & \subf{\includegraphics[width=.300\textwidth, trim={0.2cm 0.2cm 0.2cm 0.2cm},clip]{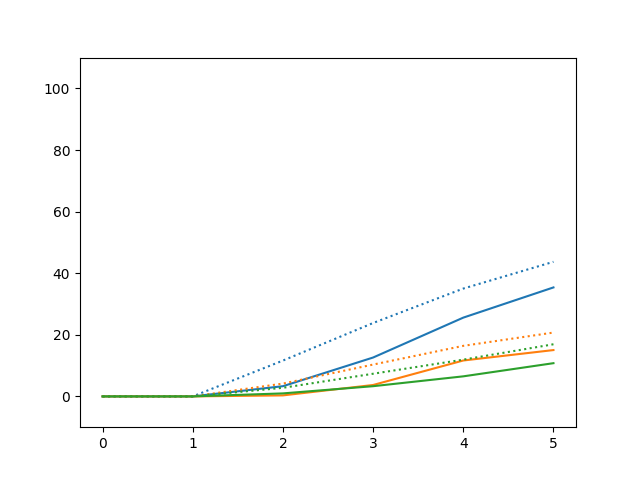}}{} & \subf{\includegraphics[width=.300\textwidth, trim={0.2cm 0.2cm 0.2cm 0.2cm},clip]{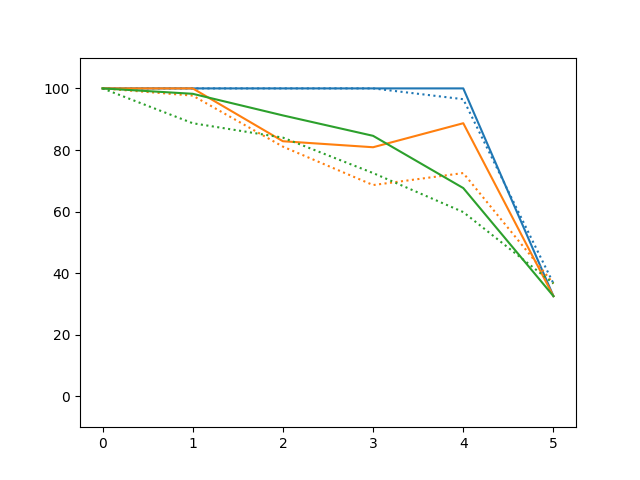}}{} & \subf{\includegraphics[width=.300\textwidth, trim={0.2cm 0.2cm 0.2cm 0.2cm},clip]{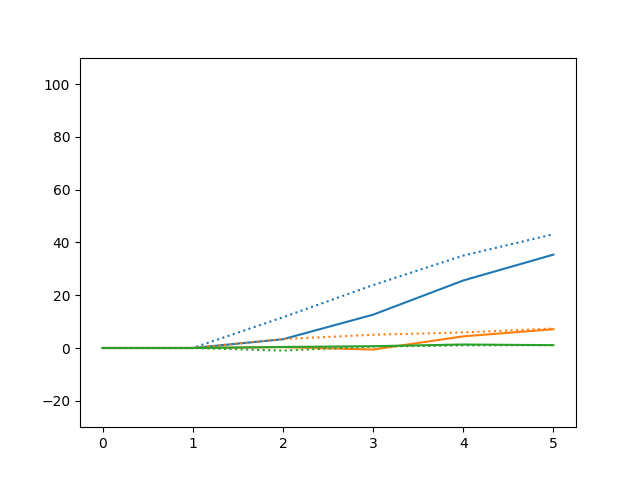}}{}\\
\subf{\includegraphics[width=0.07\textwidth ]{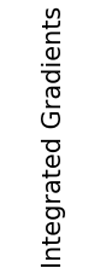}}{} & \subf{\includegraphics[width=.300\textwidth, trim={0.2cm 0.2cm 0.2cm 0.2cm},clip]{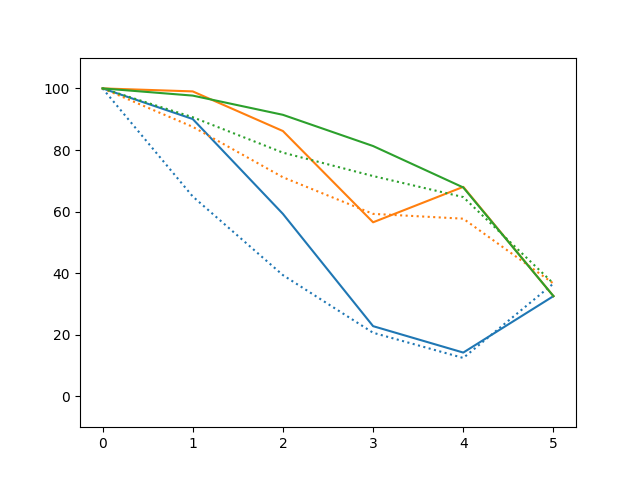}}{} & \subf{\includegraphics[width=.300\textwidth, trim={0.2cm 0.2cm 0.2cm 0.2cm},clip]{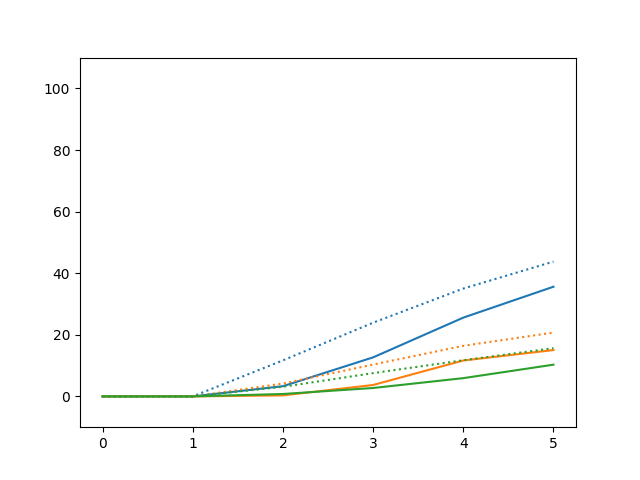}}{} & \subf{\includegraphics[width=.300\textwidth, trim={0.2cm 0.2cm 0.2cm 0.2cm},clip]{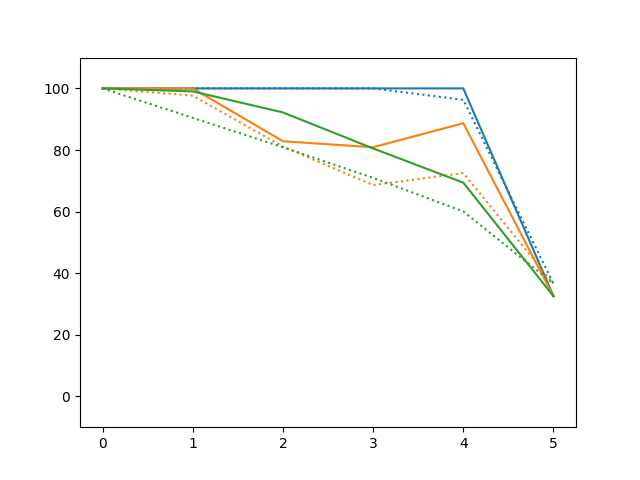}}{} & \subf{\includegraphics[width=.300\textwidth, trim={0.2cm 0.2cm 0.2cm 0.2cm},clip]{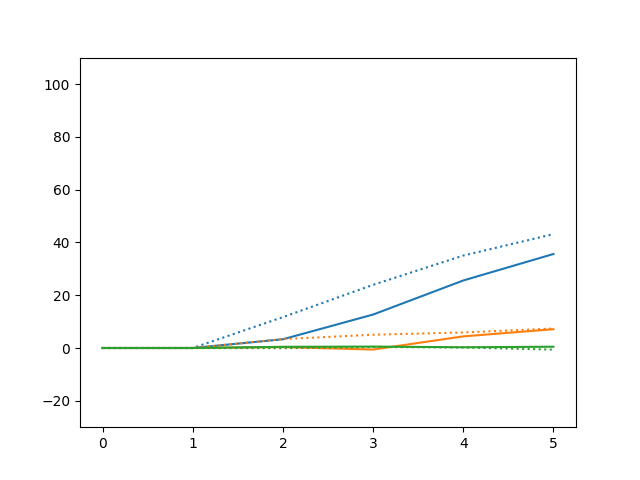}}{}\\
\subf{\includegraphics[width=0.07\textwidth ]{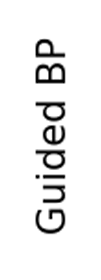}}{} & \subf{\includegraphics[width=.300\textwidth, trim={0.2cm 0.2cm 0.2cm 0.2cm},clip]{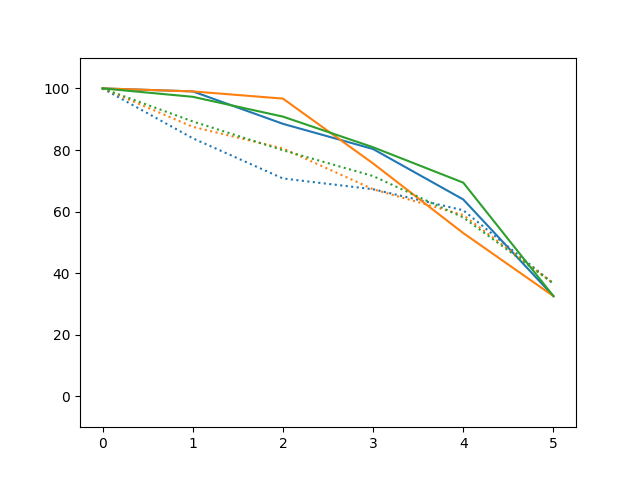}}{} & \subf{\includegraphics[width=.300\textwidth, trim={0.2cm 0.2cm 0.2cm 0.2cm},clip]{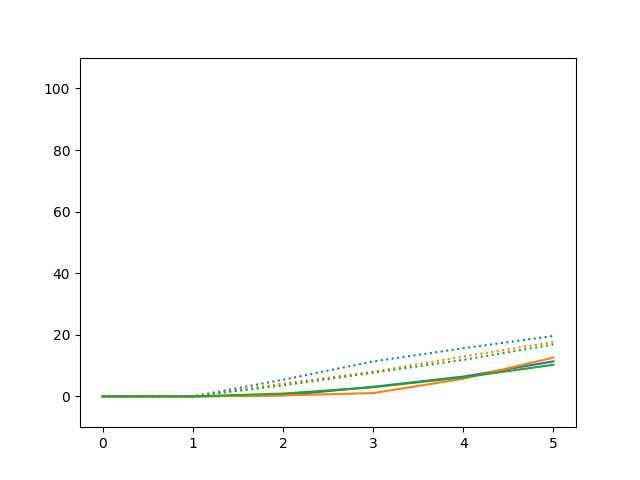}}{} & \subf{\includegraphics[width=.300\textwidth, trim={0.2cm 0.2cm 0.2cm 0.2cm},clip]{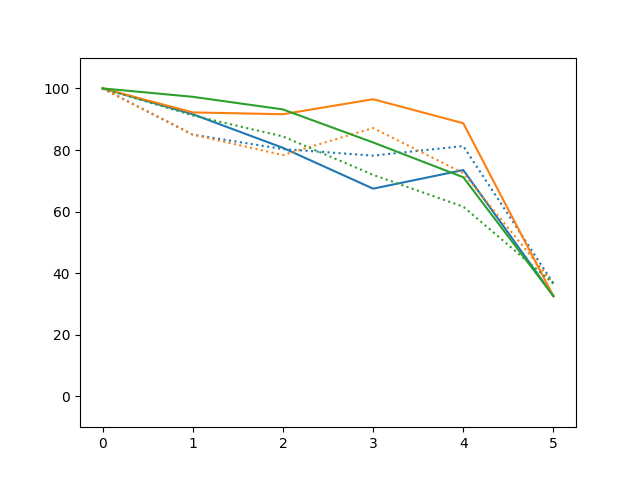}}{} & \subf{\includegraphics[width=.300\textwidth, trim={0.2cm 0.2cm 0.2cm 0.2cm},clip]{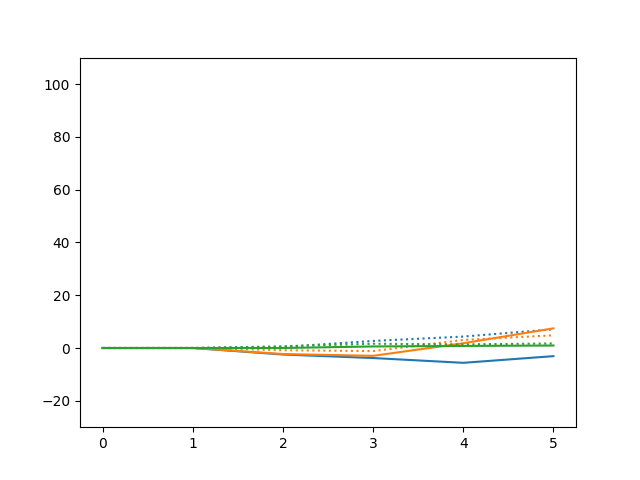}}{}\\
\subf{\includegraphics[width=0.07\textwidth ]{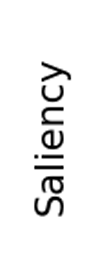}}{} & \subf{\includegraphics[width=.300\textwidth, trim={0.2cm 0.2cm 0.2cm 0.2cm},clip]{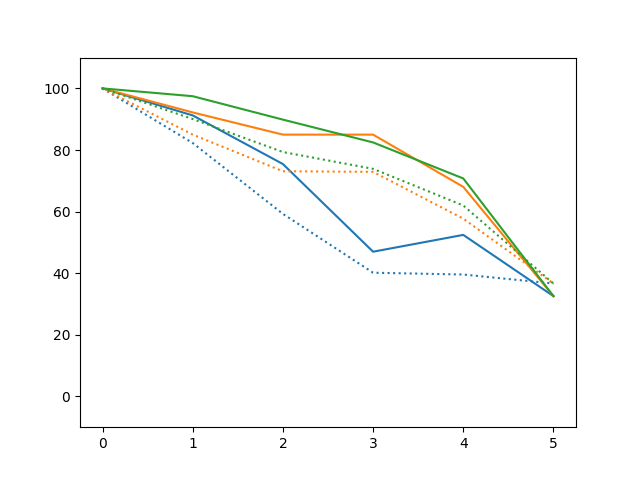}}{} & \subf{\includegraphics[width=.300\textwidth, trim={0.2cm 0.2cm 0.2cm 0.2cm},clip]{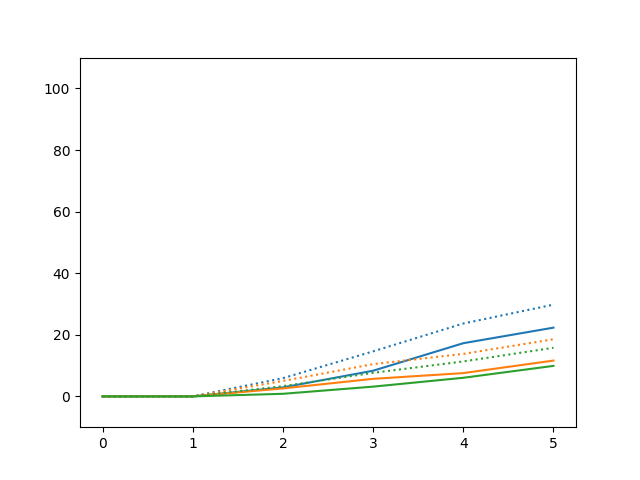}}{} & \subf{\includegraphics[width=.300\textwidth, trim={0.2cm 0.2cm 0.2cm 0.2cm},clip]{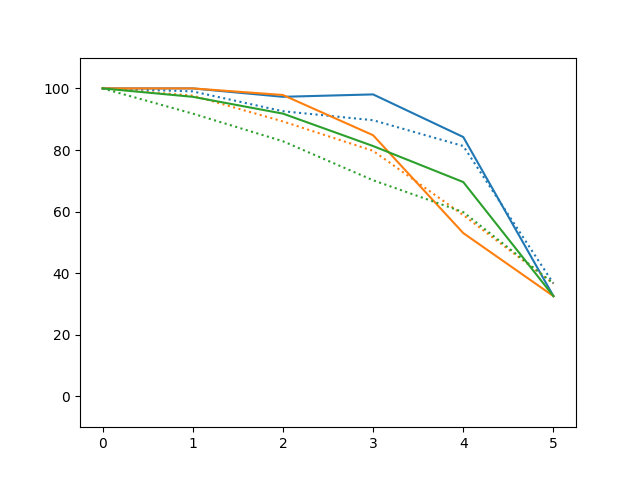}}{} & \subf{\includegraphics[width=.300\textwidth, trim={0.2cm 0.2cm 0.2cm 0.2cm},clip]{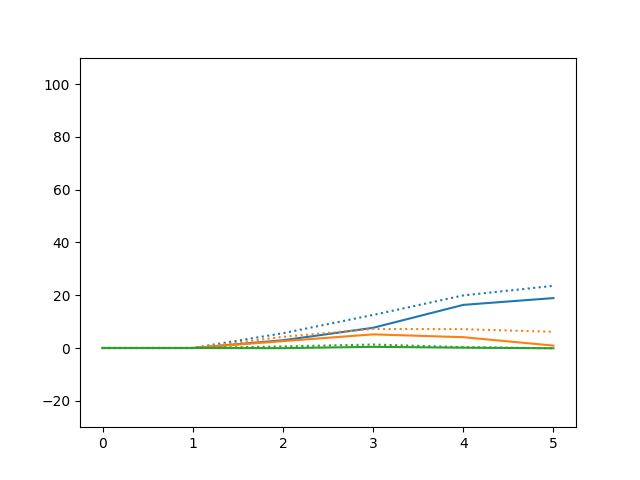}}{}\\
\subf{\includegraphics[width=0.07\textwidth ]{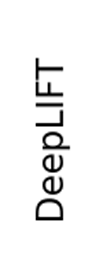}}{} & \subf{\includegraphics[width=.300\textwidth, trim={0.2cm 0.2cm 0.2cm 0.2cm},clip]{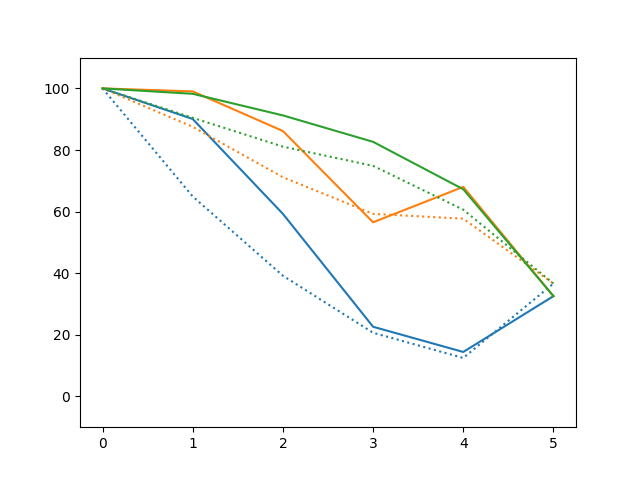}}{} & \subf{\includegraphics[width=.300\textwidth, trim={0.2cm 0.2cm 0.2cm 0.2cm},clip]{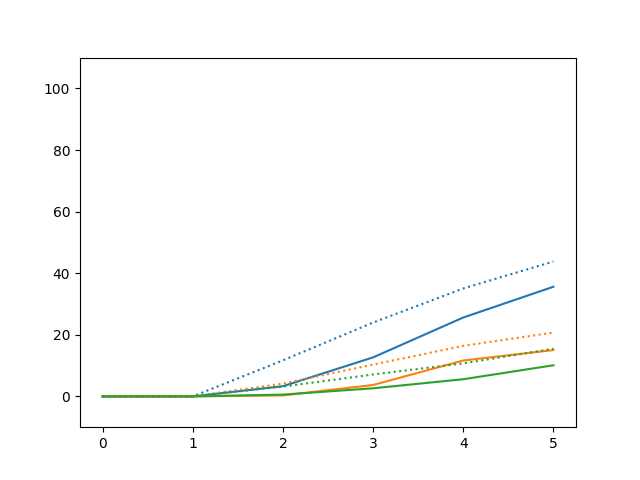}}{} & \subf{\includegraphics[width=.300\textwidth, trim={0.2cm 0.2cm 0.2cm 0.2cm},clip]{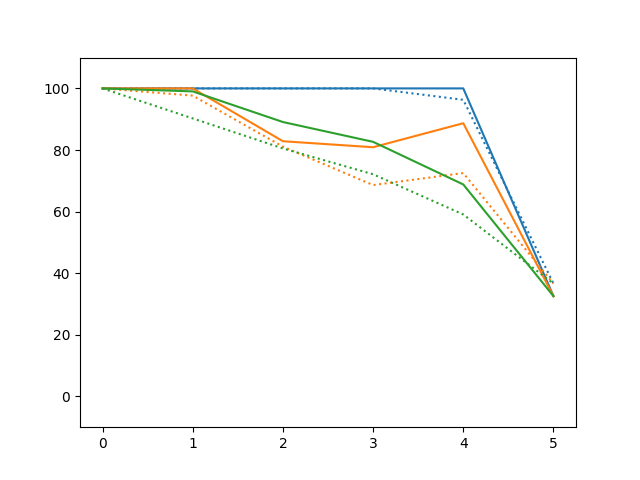}}{} & \subf{\includegraphics[width=.300\textwidth, trim={0.2cm 0.2cm 0.2cm 0.2cm},clip]{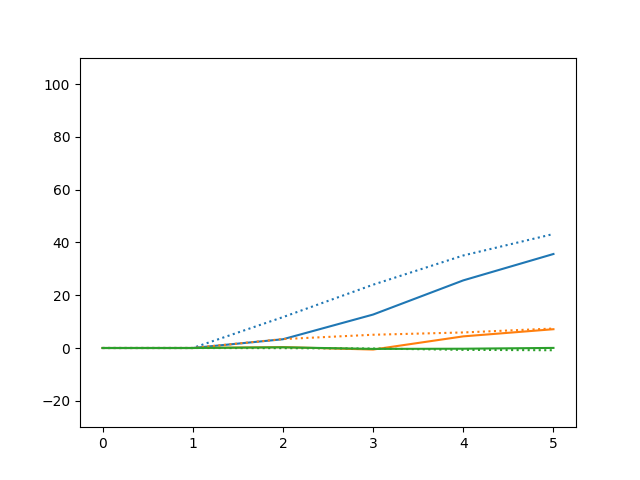}}{}\\
\hline
\end{tabular}}
\caption{MoRF (first column), AOPC (second column), LeRF (third column), and ABPC (fourth column) curves using the tested XAI methods are reported for both intra-session (solid line)  and inter-session (dotted lines) considering delta, theta, alpha, beta, gamma EEG bands  as signal components. Results scoring the input components using effective relevance (blue lines) and averaged relevance computed on training data (orange lines) are reported for each case and compared with a random component scoring (green lines). On the $x$ axis and $y$ axis are reported the iteration step in the curve generation and the accuracy level reached, respectively.}
\label{img:bands}
\end{figure*}

\setlength{\tabcolsep}{-6pt}
\begin{figure*}[p]
\scalebox{0.9}{
\begin{tabular}{ccccc}
\subf{\includegraphics[width=.07\textwidth ]{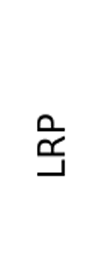}}{} & \subf{\includegraphics[width=.300\textwidth, trim={0.2cm 0.2cm 0.2cm 0.2cm},clip]{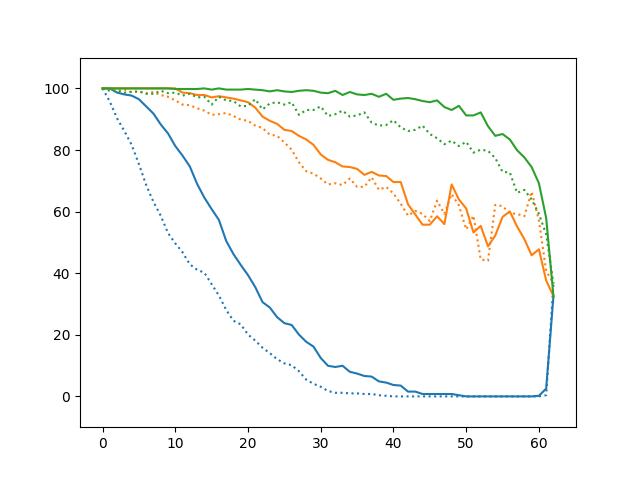}}{} & \subf{\includegraphics[width=.300\textwidth, trim={0.2cm 0.2cm 0.2cm 0.2cm},clip]{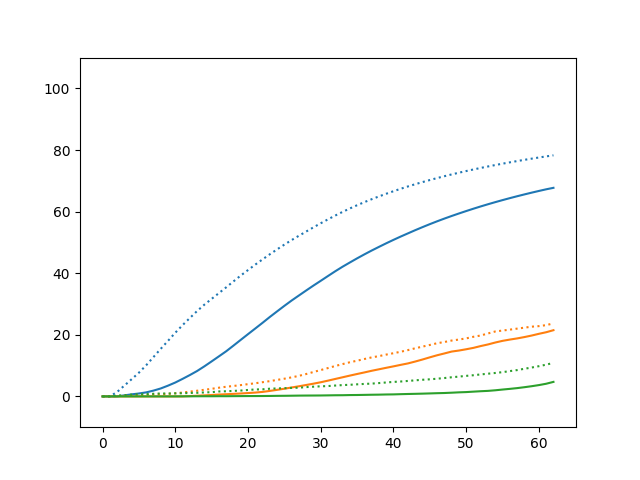}}{} & \subf{\includegraphics[width=.300\textwidth, trim={0.2cm 0.2cm 0.2cm 0.2cm},clip]{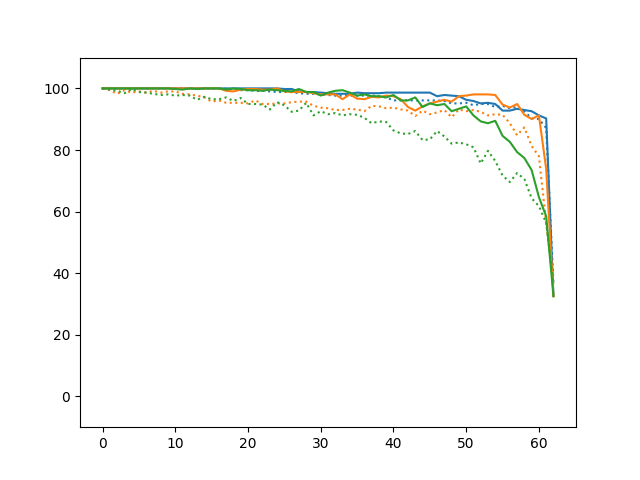}}{} & \subf{\includegraphics[width=.300\textwidth, trim={0.2cm 0.2cm 0.2cm 0.2cm},clip]{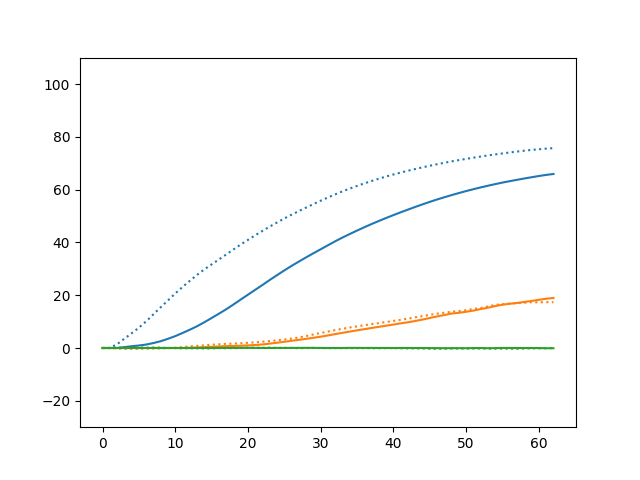}}{}\\
\subf{\includegraphics[width=.07\textwidth ]{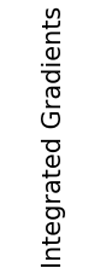}}{} & \subf{\includegraphics[width=.300\textwidth, trim={0.2cm 0.2cm 0.2cm 0.2cm},clip]{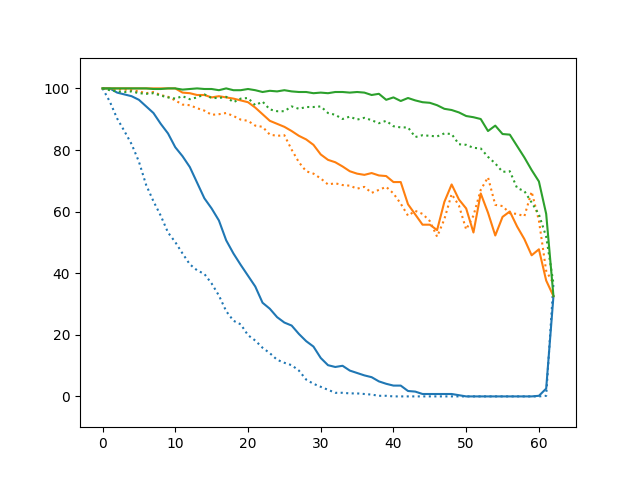}}{} & \subf{\includegraphics[width=.300\textwidth, trim={0.2cm 0.2cm 0.2cm 0.2cm},clip]{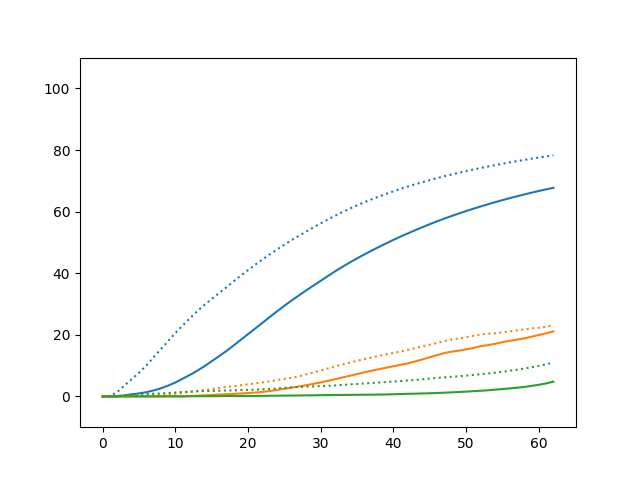}}{} & \subf{\includegraphics[width=.300\textwidth, trim={0.2cm 0.2cm 0.2cm 0.2cm},clip]{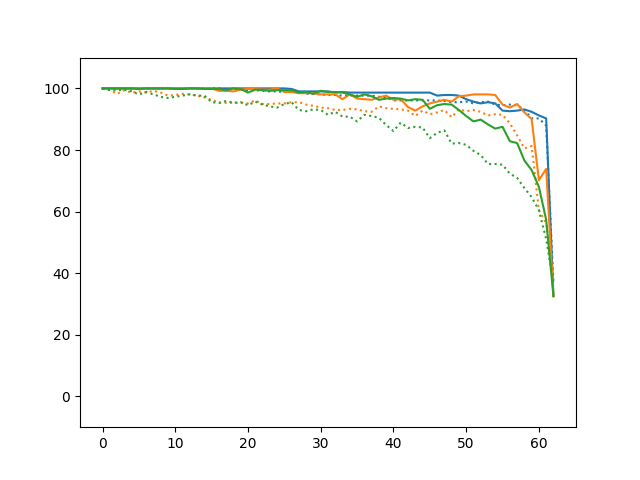}}{} & \subf{\includegraphics[width=.300\textwidth, trim={0.2cm 0.2cm 0.2cm 0.2cm},clip]{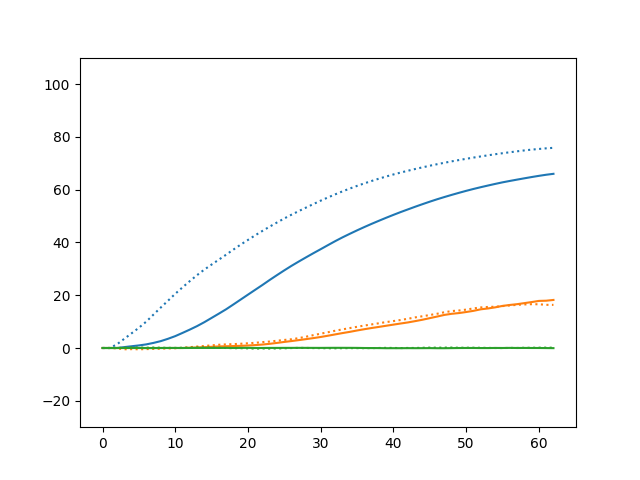}}{}\\
\subf{\includegraphics[width=.07\textwidth ]{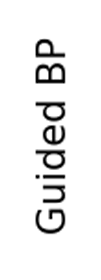}}{} & \subf{\includegraphics[width=.300\textwidth, trim={0.2cm 0.2cm 0.2cm 0.2cm},clip]{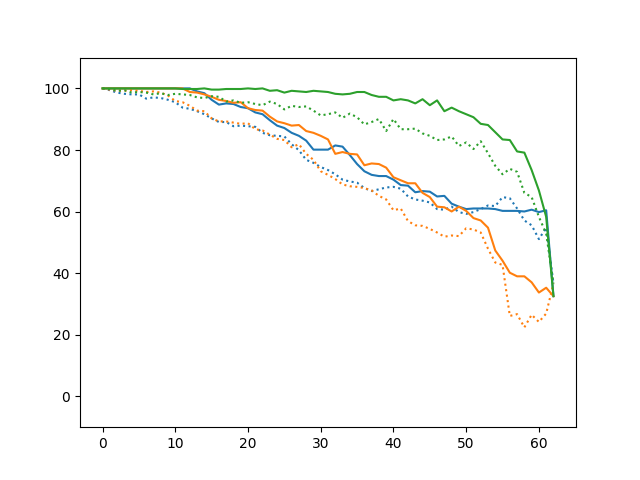}}{} & \subf{\includegraphics[width=.300\textwidth, trim={0.2cm 0.2cm 0.2cm 0.2cm},clip]{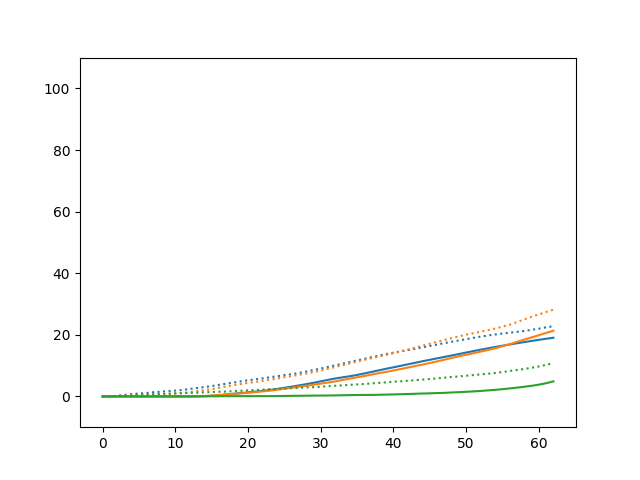}}{} & \subf{\includegraphics[width=.300\textwidth, trim={0.2cm 0.2cm 0.2cm 0.2cm},clip]{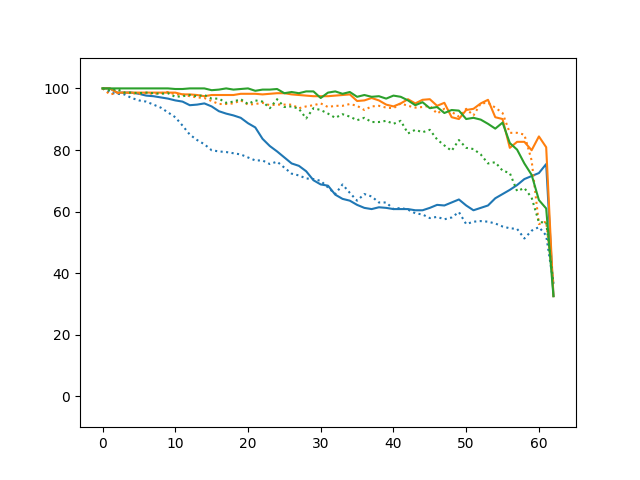}}{} & \subf{\includegraphics[width=.300\textwidth, trim={0.2cm 0.2cm 0.2cm 0.2cm},clip]{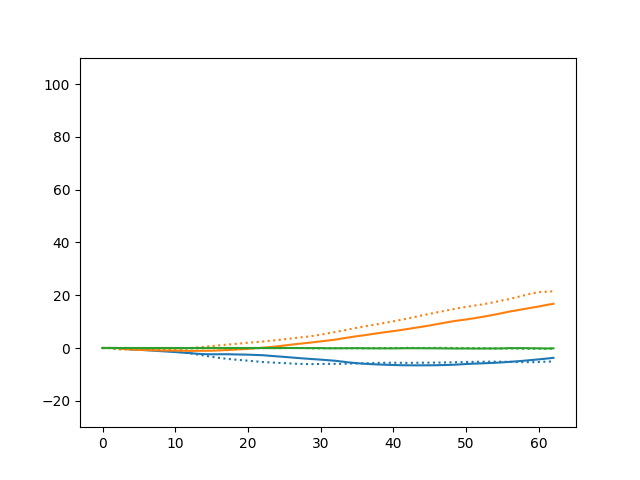}}{}\\
\subf{\includegraphics[width=.07\textwidth ]{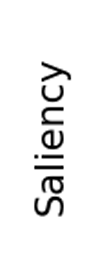}}{} & \subf{\includegraphics[width=.300\textwidth, trim={0.2cm 0.2cm 0.2cm 0.2cm},clip]{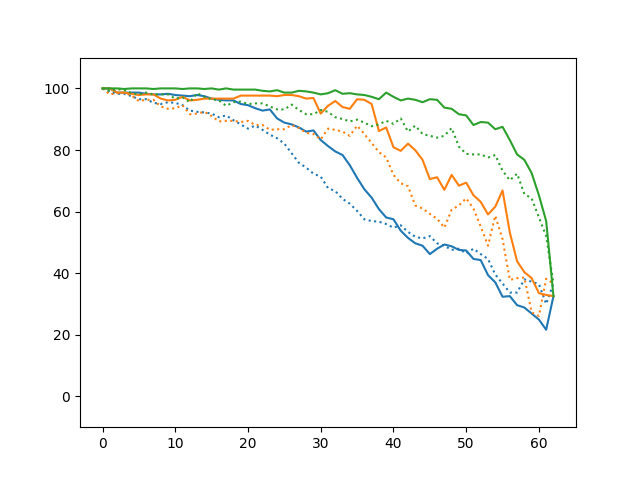}}{} & \subf{\includegraphics[width=.300\textwidth, trim={0.2cm 0.2cm 0.2cm 0.2cm},clip]{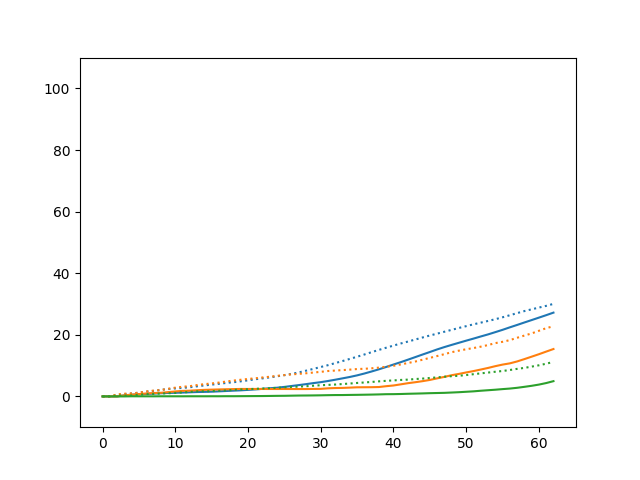}}{} & \subf{\includegraphics[width=.300\textwidth, trim={0.2cm 0.2cm 0.2cm 0.2cm},clip]{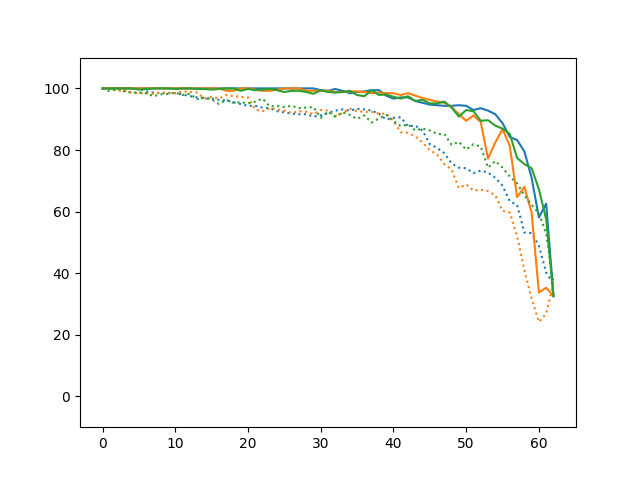}}{} & \subf{\includegraphics[width=.300\textwidth, trim={0.2cm 0.2cm 0.2cm 0.2cm},clip]{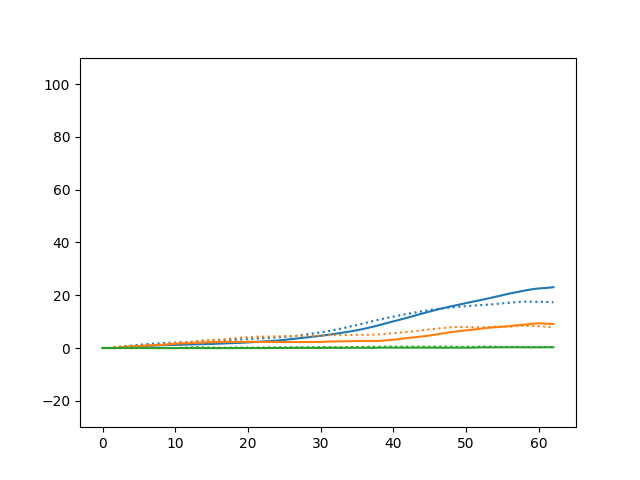}}{}\\
\subf{\includegraphics[width=.07\textwidth ]{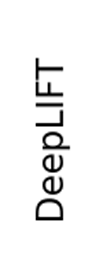}}{} & \subf{\includegraphics[width=.300\textwidth, trim={0.2cm 0.2cm 0.2cm 0.2cm},clip]{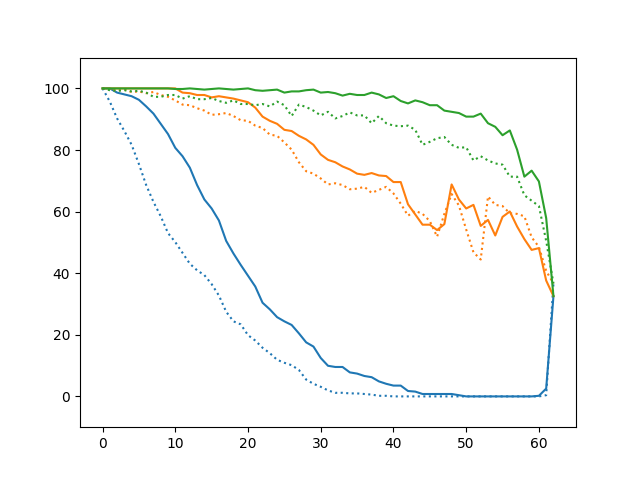}}{} & \subf{\includegraphics[width=.300\textwidth, trim={0.2cm 0.2cm 0.2cm 0.2cm},clip]{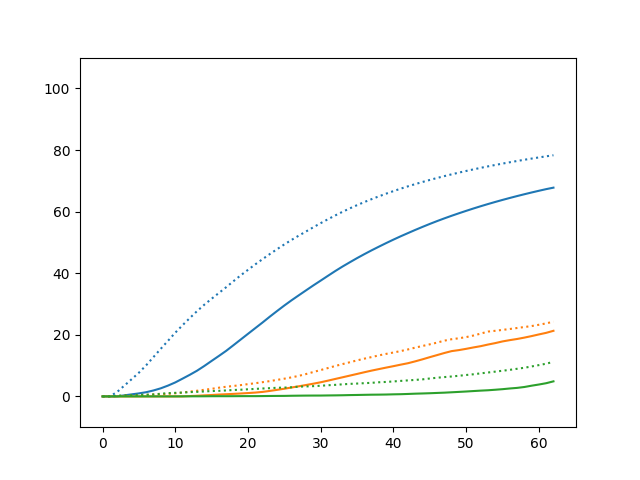}}{} & \subf{\includegraphics[width=.300\textwidth, trim={0.2cm 0.2cm 0.2cm 0.2cm},clip]{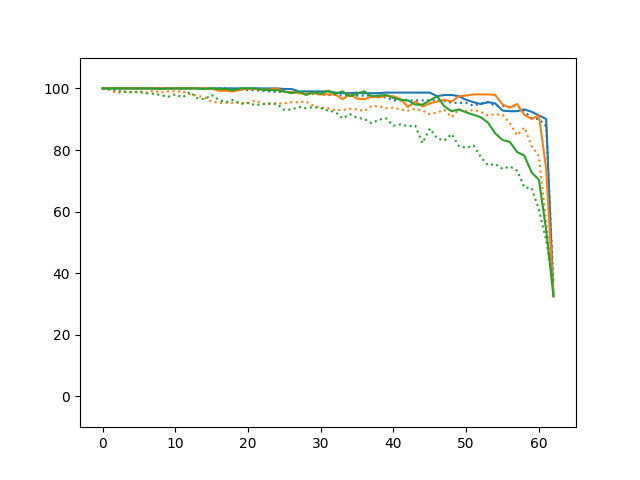}}{} & \subf{\includegraphics[width=.300\textwidth, trim={0.2cm 0.2cm 0.2cm 0.2cm},clip]{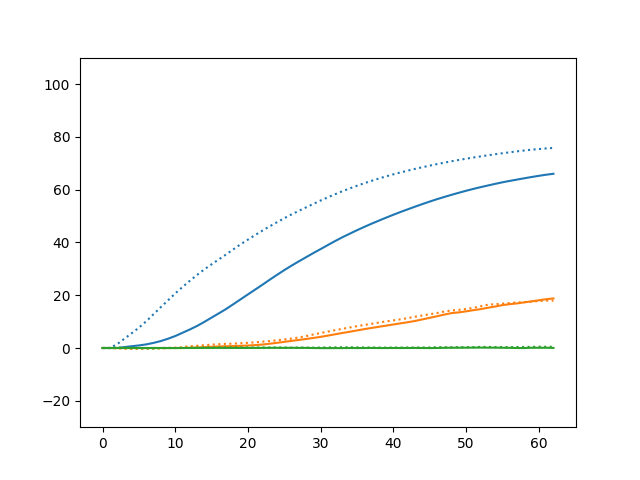}}{}\\
\hline
\end{tabular}}
\caption{MoRF (first column), AOPC (second column), LeRF (third column), and ABPC (fourth column) curves using the tested XAI methods are reported for both intra-session (solid line)  and inter-session (dotted lines) considering the acquisition electrodes as signal components. Results scoring the input components using effective relevance (blue lines) and averaged relevance computed on training data (orange lines) are reported for each case and compared with a random component scoring (green lines). On the $x$ axis and $y$ axis are reported the iteration step in the curve generation and the accuracy level reached, respectively.}
\label{img:channels}
\end{figure*}
\setlength{\tabcolsep}{-6pt}
\begin{figure*}[p]
\scalebox{0.9}{
\begin{tabular}{cccc}
\subf{\includegraphics[width=0.07\textwidth ]{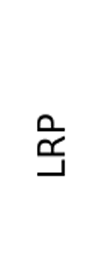}}{} &
\subf{\includegraphics[width=.350\textwidth ]{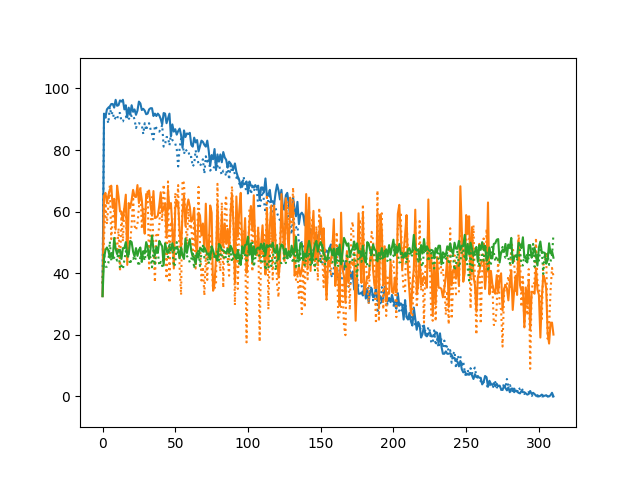}}{} & \subf{\includegraphics[width=.350\textwidth ]{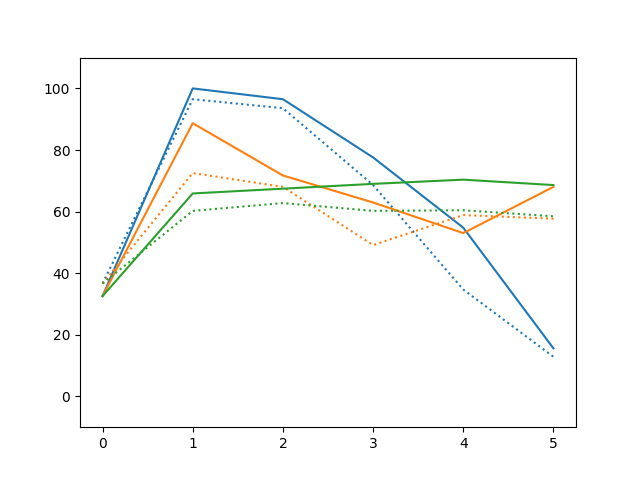}}{} & \subf{\includegraphics[width=.350\textwidth ]{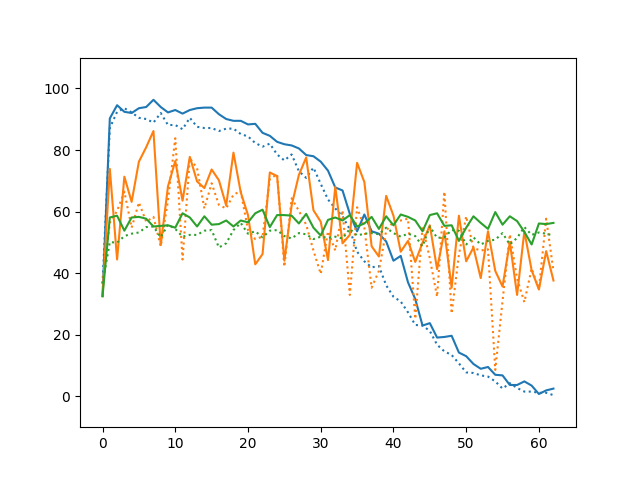}}{}\\
\subf{\includegraphics[width=0.07\textwidth ]{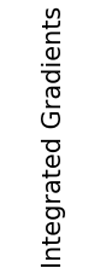}}{} & 
\subf{\includegraphics[width=.350\textwidth ]{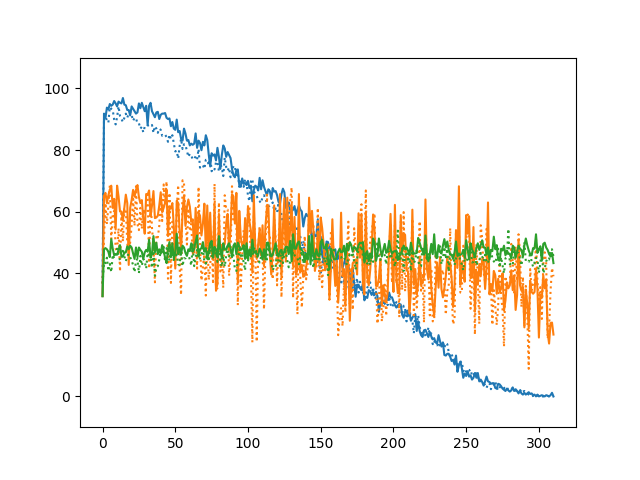}}{} & \subf{\includegraphics[width=.350\textwidth ]{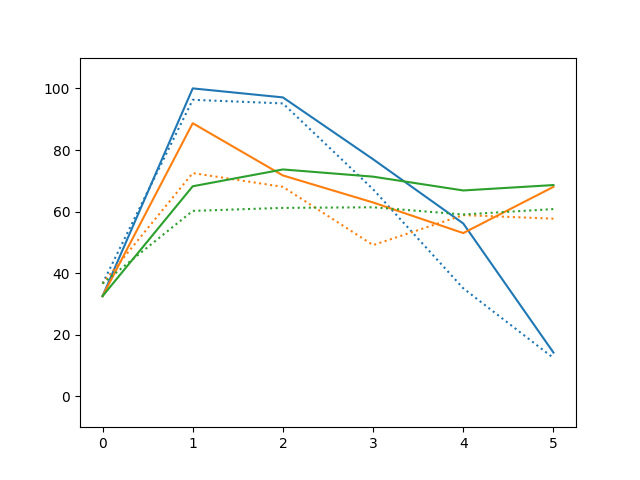}}{} & \subf{\includegraphics[width=.350\textwidth ]{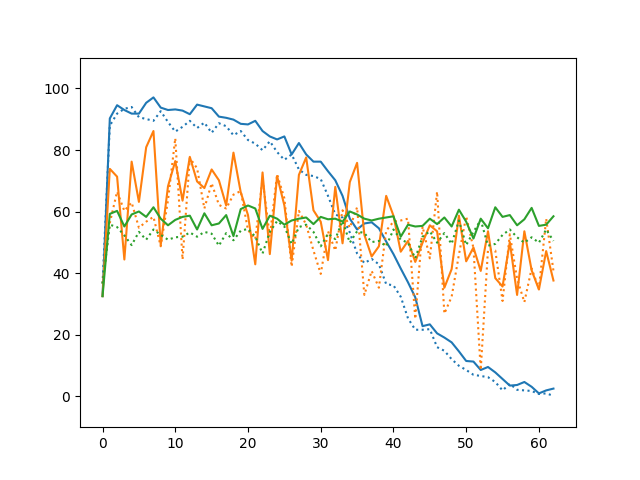}}{}\\
\subf{\includegraphics[width=0.07\textwidth ]{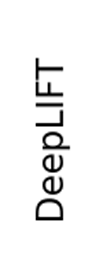}}{} &
\subf{\includegraphics[width=.350\textwidth ]{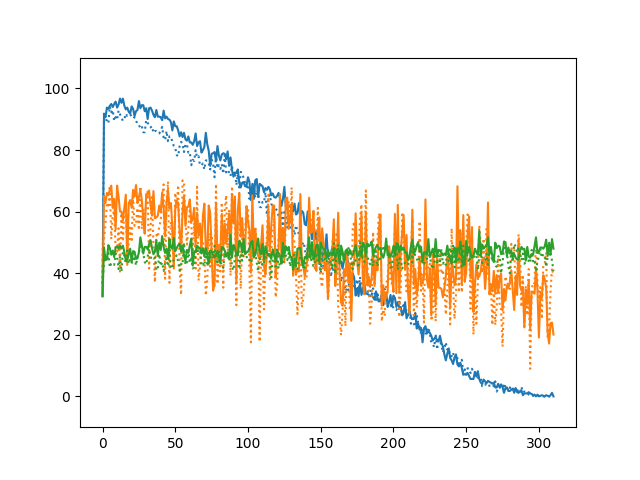}}{} &
\subf{\includegraphics[width=.350\textwidth ]{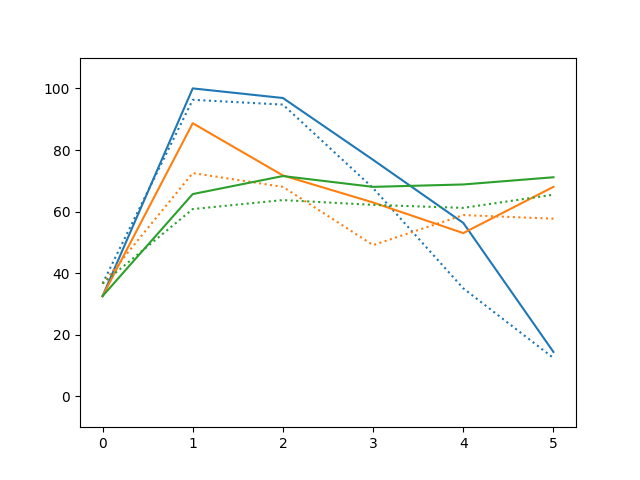}}{} & \subf{\includegraphics[width=.350\textwidth ]{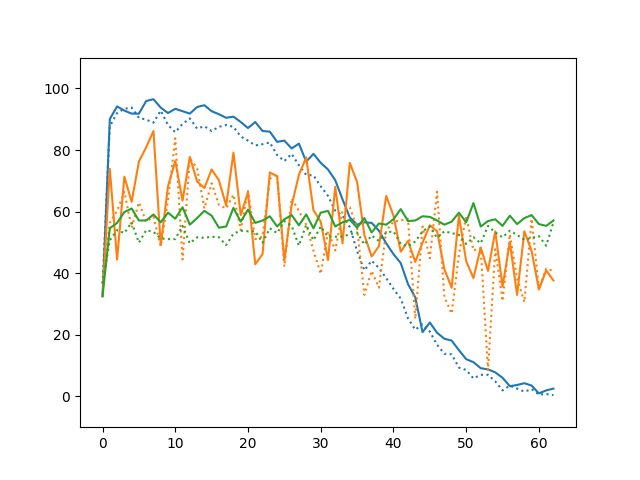}}{} \\
\hline
\end{tabular}}
\caption{
A first analysis of the discriminative power of the components alone. Signals composed of only one component following the relevance order given by the Explainer are fed to the ML system in an iterative manner. Results are reported for both intra-session (solid line)  and inter-session (dotted lines) considering features (first column), bands (second column), and electrodes (third column) as signal components. Results scoring the input components using effective relevance (blue lines) and averaged relevance computed on training data (orange lines) are reported for each case and compared with a random component scoring (green lines).}
\label{img:single}
\end{figure*}

To achieve the goals defined at the beginning of this section, the following experiments are made: firstly, to evaluate the capability of the selected XAI methods to find relevant components,  we analysed the explanations of model responses on data coming from the same session where the training data was extracted; then, to evaluate how much relevant components can be considered shared among samples of the same session, we analysed the explanations of the model responses on data belonging to a session different from the training one. %\cite{liyanage2013dynamically,meltzer2007individual}.
Finally, to evaluate if relevant components can be considered shared between samples of two different sessions and how much relevant components are dependent on the single data sample where the relevance are computed, the components' average relevance of data coming from the training session are used as sorting score and select the components belonging to another session. 

Summarising, the following cases are considered: i) intra-session case: given a model $C$ trained on data coming from a session $s_{tr}$, explanations of the responses on input data belonging to the same session $s_{tr}$ are built. ii) inter-session case: given a model $C$ trained on data coming from a session $s_{tr}$, explanation of responses on inputs belonging to a sessions $s_{te}$ different from $s_{tr}$ are built. %We consider an XAI useful for our long-term goal if it is able to highlight relevant input features both in intra-session and inter-session cases. 
Each of these cases can be in turn evaluated considering two different relevance: a) real relevance: we assume that it is possible to compute the relevance of the input, since the classification output is known; b) presumed relevance: we assume that the relevance of the input is not available, since we are outside the training stage. In this case, we use the average of the same component relevance obtained on training data as component relevance.

%Another interesting aspect is to evaluate the ability of an XAI method to "generalise", where we mean with "generalise" if the most relevant components found by an XAI method on a given sample can be considered relevant also for other unseen data. 

%To this aim, a first attempt is to evaluate how much the most relevant input components are "bound" to the owner sample, or if there are relevant features shared across sample belonging to different sessions. To give a first insight on this aspect, the adopted evaluation strategies are made considering two different cases: a) \textit{most relevant per sample}: where the relevance evaluation is made considering the relevance of each component in each single sample, and b)\textit{most relevant in average}: the evaluation is made considering the average relevance of each component  across all the sample in the set. 

\subsection{Evaluation}
For each case, we investigated the explanations returned by XAI method in order to analyse if the explanations built can correctly identify the impact that i) each input feature, ii) each electrode, and iii) each frequency band has on the classification performances. 
To this aim, we consider as relevance for each feature the relevance score returned by the XAI method, for each electrode the mean relevance score of all the feature belonging to the electrode, and for each frequency bands the mean average score of all the features belonging to the frequency band.
Therefore, the following evaluation strategies are then adopted and repeated considering features, electrodes, and frequency bands as EEG components in turn: a) analysis of the MoRF (Most Relevant First) curve, proposed in \cite{bach2015pixel, samek2016evaluating}. In case of evaluating the components relevance returned by the explanation method, the MoRF curve can be computed as follows: given a classifier, an input EEG signal $\mathbf{x}$ and the respective classification output $C(\mathbf{x})$, the EEG components are iteratively replaced by zeros, following the descending order with respect to the relevance values returned by the explanation method. In other words, performances were analysed by removing (i.e. setting to zero) components in a decreasing order of impact on the predictions supplied by the explanation. In this way, the expected curve is such that more relevant the identified components are for the classification output, steepest is the curve. Furthermore, the change in the AOPC (Area Over Perturbation Curve) value is reported for each MoRF iteration. AOPC is computed as $$AOPC=\frac{1}{K+1}\langle \sum\limits_{k=0}^K C(\mathbf{x}^{(0)})-C(\mathbf{x}^{(k)})) \rangle$$ where $K$ is the total number of iterations, $\mathbf{x}^{(0)}$ is the original input, $\mathbf{x}^{(k)}$ is the input at the iteration $k$, and $\langle \cdot \rangle$ is the average operator over a set of inputs. MoRFs and AOPCs are reported also considering channels and bands as characteristics to analyse.

b) the analysis of the LeRF (Least Relevant First) curve, proposed in \cite{samek2016evaluating}. Differently from the MoRF curve, in this case the EEG components are iteratively removed following the ascending order with respect to the relevance values returned by the explanation method. In the resulting curve, we expect that the classification output should be very close to the original value  when the less relevant components are removed (corresponding to the first iterations), dropping quickly to zero as the process goes toward the remotion of relevant elements. While the MoRFs report how much the classifier output is destroyed removing highly relevant components, LeRFs report how much the least relevant components leave the output intact. These indications can be combined in the ABPC (Area Between Perturbation Curves, \cite{samek2016evaluating}) quantity, defined as:
$$ABPC=\frac{1}{K+1}\langle \sum\limits_{k=0}^K C(\mathbf{x}_{MoRF}^{(k)})-C(\mathbf{x}_{LeRF}^{(k)})) \rangle$$
where $\mathbf{x}_{MoRF}^{(k)}$, $\mathbf{x}_{LeRF}^{(k)}$ are the values of the MoRF and LeRF values obtained at the $k$-th iteration step. ABPC is an indicator of how
good the XAI method is. The larger the ABPC value, the better the XAI method. LeRFs and ABPCs are reported also for channels and bands analysis.

c) an analysis of the discriminative power of each component alone is made. Signals composed of only one component following the relevance order
given by the XAI method are fed to the ML system in an iterative manner, and the relative performance curves are plotted. 

All the experiments were carried out only on correctly classified samples.
%Finally, the per-sample relevance and the average relevance over all the training set were used for each evaluation case.

\subsection{Classification model}

The XAI methods are evaluated on a feed-forward fully connected multi layered neural networks.
Hyperparameters were tuned through bayesian optimisation \cite{snoek2012practical}: the number of layers was constrained to a maximum of 3; for each layer, the number of nodes was searched in the space $\{2^n| n \in \{4, 5, ..., 10\}\}$ having the ReLU as activation function.
Each experiment was run having early stopping as convergence criterion with 20 epochs of patience. The 10 \% of the training set was extracted using stratified sampling \cite{neyman1992two} on class labels and considered as validation set. Network optimisation was performed using Adam optimiser \cite{kingma2014adam}, whose learning rate that was searched in the space $\{0.1, 0.01, ..., 0.0001\}$.

As a result from the model selection stage, the best setting consisted in ANN having 3 layers with 128, 256 and 128 neurons respectively. The learning rate was set to $0.01$, and reduced to its 10 \% whenever the loss on validation set plateaus for 10 consecutive epochs. 

% Explainability of models were produced following several methods: Layer-wise Relevance Propagation (LRP) \cite{bach2015pixel}, Integrated Gradients \cite{sundararajan2017axiomatic}, Saliency \cite{simonyan2013deep}, DeepLIFT \cite{shrikumar2017learning} and Guided Backpropagation \cite{springenberg2014striving}.

\section{Results \& discussions}
\label{sec:results}
Since the behaviour of the explored XAI methods resulted in being similar across all the subjects, we report only the results obtained on just one subject. 
In Fig. \ref{img:feature}, \ref{img:bands}, and \ref{img:channels} MoRF and LeRF curves using the tested XAI methods are reported for both intra-session and inter-session cases, considering as components to remove at each step features (Fig. \ref{img:feature}), bands (Fig. \ref{img:bands}), and channels (Fig. \ref{img:channels}), respectively.
Results related to the intra-session cases are reported with solid lines, while those regarding the inter-session case are marked with dotted lines. On the $x$ axis and $y$ axis are reported the iteration step in the curve generation and the accuracy level reached, respectively. With blue lines, results scoring the input components using effective relevance are reported; with orange lines, results scoring the components using averaged relevance computed on training data are reported; with green lines, results related to random choice.

All the curves were compared with the random curve obtained by removing the components in random order. Several interesting points can be highlighted:

1) In all the cases, LRP, IG and Deep LIFT resulted in being more reliable XAI methods with respect to Saliency and Guided BP. Indeed, MoRF curves of LRP, IG and Deep LIFT have high slopes, however similar to each other, differently from Saliency and Guided BP. In particular, the latter is the only method among those tested whose explanations do not always seem to capture the relevant components, especially in the case of intra-session. These considerations seem consistent with what is reported in LeRF, AOPC, and ABPC.

2) counterintuitively, in almost all the cases, explanations built in inter-session cases seem to be more reliable with respect to intra-session cases. This behaviour can be explained by a more significant "robustness" of the trained classifier toward data from the same training session. Instead, data coming from different sessions leads the classifier toward more borderline class scores, and minimum perturbation of the input data can lead to different classes, influencing the final performance.

3) Although the best XAI methods can locate relevant features/channels/bands for each input data sample, they don't seem able to locate a set of relevant components for all the samples. In other words, the examined XAI methods fail to "generalise" to a set of general features/channels/bands relevant to the most significant part of the possible inputs. Indeed, removing the components following the average relevance (obtained in the training stage) in reverse order (MoRF orange curves) does not lead to a steep drop in performance, as in the other case (MORF blue curves). Even in some cases, such as using bands as a component to assign the relevance (Fig. \ref{img:bands}), the obtained curves overlap with the random ones, highlighting that removing bands in random order is almost the same that following the relevance assigned by the XAI method. This is confirmed by the other evaluation metrics adopted, i.e. MeRF, AOPC and ABPC curves.

In Fig. \ref{img:single} a first analysis of the discriminative power of the components alone is made. 
Signals composed of only one component following the relevance order given by the XAI method are iteratively fed to the ML system. We limit the analysis only to the best XAI methods identified in the previous step: DeepLIFT, IG and LRP. From the obtained results, it is interesting to notice that the components considered most relevant for each sample fed to the classifier are enough to reach high performances. However, considering the average relevance detected during the training stage, the best components do not seem to lead toward similar performance, although they are still better than a random choice.

\section{Conclusions}
\label{sec:conclusions}
In this work, the performances of several XAI methods proposed in the literature in the context of Brain-Computer Interface (BCI) problems using EEG input-based Machine Learning (ML) algorithms are experimentally evaluated.
The focus was on how much the relevant components selected by XAI methods be shared between different samples of the same dataset (in this case, same session)  or samples of different datasets (in this case, different sessions).  
The final results show that the components considered most relevant for each sample fed to the classifier are enough to achieve high performances. However, the components detected considering the best average relevance during the training stage do not seem to lead toward performance returned by components scored according to their effective relevance returned by the XAI method.

This work is the first step toward developing a BCI system able to exploit XAI methods to alleviate the dataset shift problem. However, in this work, only data belonging to different sessions but acquired from the same subjects are taken into account. In future work, we plan to analyse the behaviour of XAI methods with inter-subject classifiers.
Several benefits can be obtained in the EEG-based BCI applications by the proposed project. For example, a BCI system can work across different subjects without retraining the model on each new unseen subject (subject-independent model).
Furthermore, a better understanding of the relationships between the system inputs and outputs provided by XAI explanations can lead to the developing and producing more effective EEG acquisition devices.

\section*{acknowledgments}
This work is supported by the European Union - FSE-REACT-EU, PON Research and Innovation 2014-2020 DM1062/2021 contract number 18-I-15350-2 and by the Ministry of University and Research, PRIN research project "BRIO – BIAS, RISK, OPACITY in AI: design, verification and development of Trustworthy AI.", Project no. 2020SSKZ7R .
\bibliographystyle{plain}
\bibliography{_bib}
\end{document}